\documentclass{article}
\usepackage[main, final]{neurips_2026}
\usepackage[utf8]{inputenc}
\usepackage[T1]{fontenc}
\usepackage{hyperref}
\usepackage{url}
\usepackage{booktabs}
\usepackage{amsfonts}
\usepackage{amsmath}
\usepackage{amssymb}
\usepackage{nicefrac}
\usepackage{microtype}
\usepackage[table]{xcolor}
\definecolor{tabhighlight}{RGB}{232,245,255}
\definecolor{closedmodel}{RGB}{230,230,230}
\usepackage{graphicx}
\usepackage{subcaption}
\usepackage{multirow}
\usepackage{makecell}
\usepackage{enumitem}
\usepackage{wrapfig}
\usepackage{tabularx}
\usepackage{arydshln}
\newcommand{\method}{VPE}
\newcommand{\methodfull}{Visual Prompt Engineering}

\newif\ifshowchanges
\showchangestrue

\usepackage[normalem]{ulem}

\usepackage{algorithm}
\usepackage{algpseudocode}

\title{Imagine Before You Draw: \methodfull{} for Image Generation}

\author{%
  Liyu Jia$^{1,*}$ \;
  Fengda Zhang$^{1,*,\dagger}$ \;
  Jiachun Pan$^{2,*}$ \;
  Kesen Zhao$^{1}$ \;
  Saining Zhang$^{1}$ \\
  \bfseries Wang Lin$^{3}$ \;
  Weijia Wu$^{2}$ \;
  Yue Liao$^{2}$ \;
  Aojun Zhou$^{4}$ \;
  Hanwang Zhang$^{1}$ \\[4pt]
  \mdseries $^{1}$Nanyang Technological University \;
  $^{2}$National University of Singapore \\
  $^{3}$Zhejiang University \;
  $^{4}$The Chinese University of Hong Kong \\[2pt]
  \texttt{ji0011yu@e.ntu.edu.sg} \; \texttt{fdzhang328@gmail.com}
}

\begin{document}
\maketitle

\begin{abstract}
Incorporating visual semantic representations as an intermediate step before image generation can reduce the modeling difficulty between text and images, thereby improving generation quality.
Recent works such as X-Omni and BLIP3o-Next have explored this direction, but they typically use a two-stage external pipeline: a separate autoregressive model first generates semantic tokens, which are then fed as conditioning to an independent diffusion decoder. When the decoder is conditioned only on the compressed semantic plan, it cannot directly access the original input, which introduces an information bottleneck that limits detail preservation in downstream tasks such as editing.
Internal architectures such as Transfusion, BAGEL, and Show-o2 avoid this bottleneck by enabling cross-modal interaction within a single model, but they still face the difficult text-to-pixel modeling gap without intermediate semantic guidance.
We propose \methodfull{} (\method{}), which can be seamlessly integrated into such internal frameworks. Specifically, the model first autoregressively generates visual semantic tokens (\textit{e.g.}, SigLIP 2) as ``visual prompts'' that capture the semantic layout, then generates the full image tokens conditioned on this plan.
We validate \method{} across class-conditional generation, text-to-image generation, and image editing, covering various token types and model architectures. Results show that \method{} can accelerate convergence, raise quality ceilings, and through internal integration, achieve substantially better editing preservation (PSNR: 26.76 vs.\ 19.92) than a matched external pipeline with semantic-only decoder conditioning, while maintaining competitive editing responsiveness.
\end{abstract}

\section{Introduction}
\label{sec:intro}

Recent multimodal models~\cite{showo,showo2,emu3,transfusion,bagel} have made rapid progress in image generation.
A central challenge in these models is bridging the gap between high-level semantic conditioning (\textit{e.g.}, class labels or text prompts) and low-level pixel-accurate image synthesis.
Incorporating visual semantic representations as intermediate signals has emerged as a key trend, as it can reduce the modeling difficulty and improve generation quality.

Several recent works have begun exploring this direction. X-Omni~\cite{xomni} and BLIP3o-Next~\cite{blip3o} employ SigLIP 2~\cite{siglip2} tokens as conditioning signals for image generation. However, they adopt an \textit{external} architecture (Figure~\ref{fig:main}b) where a separate autoregressive model produces SigLIP 2 tokens that are then fed as conditioning to an independent diffusion model. When the diffusion decoder is conditioned only on the compressed semantic plan, it cannot directly access the original conditioning, creating an information bottleneck that is particularly harmful for tasks requiring detail preservation such as image editing. In contrast, \textit{internal} architectures (Figure~\ref{fig:main}a) such as Transfusion~\cite{transfusion}, BAGEL~\cite{bagel}, and Show-o2~\cite{showo2} avoid this bottleneck by enabling cross-modal interaction within a single model, but they have not incorporated visual semantic representations, leaving the difficult text-to-pixel modeling gap unaddressed.
In addition, no prior work has systematically investigated whether visual semantic prompts benefit different types of image tokens (discrete tokens such as Emu3~\cite{emu3} and continuous tokens such as VAE vectors~\cite{vqvae}), how they affect convergence and quality ceilings, or what are the trade-offs between these two architectural paradigms.

Drawing an analogy from chain-of-thought reasoning in large language models~\cite{cot}, where decomposing complex reasoning into intermediate steps dramatically improves performance, we propose \textit{\methodfull{}} (\method{}) to address these challenges. \method{} inserts a compact set of SigLIP 2 visual tokens that capture the high-level semantic layout of the target image before the model generates the full image tokens, transforming a single difficult generation step into two easier sub-problems: \textit{semantic planning} (what to draw) followed by \textit{detail rendering} (how to draw it). 
However, learning reliable semantic plans remains non-trivial. If the model is naively conditioned on ground-truth SigLIP 2 visual prompts during training, it can over-rely on these highly informative tokens and pay insufficient attention to other conditioning signals. At inference time, however, the model must condition on its own predicted visual prompts, which are inevitably imperfect; this train--inference discrepancy can lead to severe quality degradation. This issue is further amplified in continuous-token models, where the cross-entropy loss for predicting SigLIP 2 tokens and the flow-matching loss for image tokens operate at different scales, making joint optimization unstable.
To address these challenges, we introduce a progressive training schedule that gradually increases the model's reliance on visual prompts (Section~\ref{sec:schedule}). 
Crucially, \method{} supports both internal integration (shared attention within a single model) and external integration (separate AR + DiT), allowing us to systematically analyze these paradigms.

We validate \method{} across class-conditional generation, text-to-image generation, and image editing, showing consistent improvements across these tasks.
Our contributions are threefold:
\begin{itemize}[leftmargin=*,itemsep=0pt, topsep=0pt]
\item We propose \method{}, a general technique for improving image generation by inserting SigLIP 2 visual tokens into internal architectures. Through experiments, we validate its effectiveness across discrete and continuous token types, various image generation tasks, and multiple model architectures including a controlled comparison of internal and external paradigms.
\item We identify the train-inference gap and loss-scale imbalance as key challenges in \method{} training, and provide a progressive training schedule that effectively addresses both issues for autoregressive and diffusion objectives.
\item Our experimental results show that internal architectures preserve finer details in editing than external ones with semantic-only decoder conditioning, but directly modeling text-to-image within a single model is inherently harder to converge. \method{} alleviates this difficulty by providing intermediate semantic guidance, accelerating the convergence of internal architectures while retaining their editing advantage.
\end{itemize}

\section{Method}
\label{sec:method}

\subsection{Visual Prompt Engineering}
\label{sec:vpe}

Given a conditioning signal $c$ (class label, text prompt, or reference image) and a target image $I$, standard image generation models learn the mapping $c \to \mathbf{T}_{\text{img}}$, where $\mathbf{T}_{\text{img}}$ represents the image tokens (discrete codebook indices or continuous latent vectors). \method{} introduces an intermediate semantic representation by first generating \textit{visual prompts} $\mathbf{T}_{\text{vp}}$ (SigLIP 2~\cite{siglip2} tokens extracted from the target image) before generating the image tokens:
\begin{equation}
c \;\to\; \underbrace{\mathbf{T}_{\text{vp}}}_{\text{semantic plan}} \;\to\; \underbrace{\mathbf{T}_{\text{img}}}_{\text{detail rendering}}
\label{eq:vpe}
\end{equation}

During training, the visual prompts are obtained by encoding the ground-truth image $I$ with a frozen visual semantic encoder, specifically SigLIP 2-Giant-Patch16-384~\cite{siglip2,xomni} in this paper. With a patch size of 16, the encoder produces a compact token grid at $16\times$ spatial downsampling: for example, $384\times384 \to 576$ tokens and $224\times224 \to 196$ tokens. The resulting tokens are then discretized using the frozen SigLIP-VQ tokenizer from X-Omni~\cite{xomni}, which adopts a codebook of 16,384 entries. At inference time, the model autoregressively generates $\mathbf{T}_{\text{vp}}$ from $c$, then conditions image token generation on both $c$ and the self-generated $\mathbf{T}_{\text{vp}}$. SigLIP 2 tokens capture high-level semantic content (object categories, spatial layout, color distribution) while abstracting away pixel-level details, making them an effective intermediate representation for semantic planning.

\paragraph{Compatibility with different token types.}
A key advantage of \method{} is its generality. The visual prompt generation is always autoregressive (predicting discrete SigLIP 2 token indices via cross-entropy loss), while the subsequent image generation can use \textit{any} framework. For \textbf{discrete} image tokens such as Emu3~\cite{emu3}, the full sequence is $[c, \mathbf{T}_{\text{vp}}, \mathbf{T}_{\text{img}}]$, trained end-to-end with cross-entropy loss. For \textbf{continuous} image tokens such as VAE vectors, the image is generated via diffusion or flow matching~\cite{flowmatching} conditioned on the visual prompts, using cross-entropy loss for $\mathbf{T}_{\text{vp}}$ and the corresponding continuous generation loss for $\mathbf{T}_{\text{img}}$.
\method{} only adds SigLIP 2 vocabulary embeddings to the model, resulting in $<4\%$ additional parameters in our experiments.

\begin{figure}[t]
\centering
\includegraphics[width=\linewidth]{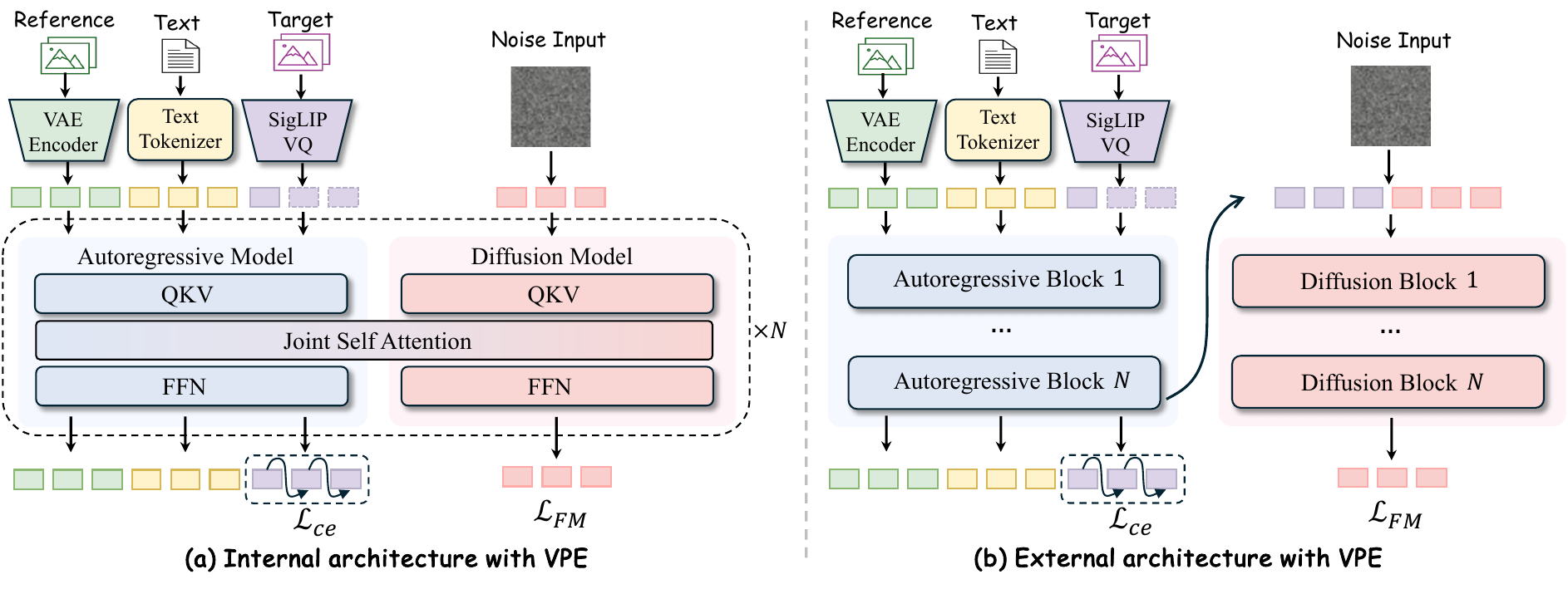}
\caption{\textbf{\method{} in two architectural frameworks.} \method{} inserts SigLIP 2 visual prompts before image generation as a semantic plan. \textbf{(a)} In the internal architecture, all modalities (text, SigLIP 2 tokens, image tokens) share attention within a single model via Eq.~\ref{eq:mot_attn}, enabling direct cross-referencing during generation. \textbf{(b)} In the external architecture, an AR model first produces SigLIP 2 features, which are then routed to a separate DiT decoder. While both frameworks benefit from \method{} for generation, the external pipeline with semantic-only DiT conditioning introduces an information bottleneck that limits editing fidelity (Section~\ref{sec:theory}).}
\label{fig:main}
\end{figure}

\subsection{Progressive Training Schedule}
\label{sec:schedule}

In our experiments, we observe that if the model is trained with ground-truth visual prompts from the start, it becomes overly dependent on them. At inference time, the model must rely on its own autoregressively generated SigLIP 2 tokens, which are inevitably imperfect in early training. This train-inference discrepancy causes significant quality degradation. To ensure the model can still generate reasonable images even when the visual prompts are noisy or incorrect, we introduce a progressive schedule that gradually increases the model's reliance on visual prompts, preventing it from degenerating into a simple copying task that only works with perfect SigLIP 2 conditioning. Additionally, for continuous token models, the cross-entropy loss for SigLIP 2 tokens and the flow matching loss for image tokens operate at different scales. Without careful balancing, the cross-entropy gradient can dominate and impede the learning of flow matching denoising.

\paragraph{For discrete token models.}
We define a step-dependent masking probability:
\begin{equation}
p_{\text{mask}}(s) = p_0 \cdot (1 - \sigma(s)) + p_1 \cdot \sigma(s), \quad \sigma(s) = \frac{1}{1 + e^{-k(s/S - 0.5)}}
\label{eq:schedule}
\end{equation}
where $s$ is the current step, $S$ is total steps, $p_0{=}0.95$, $p_1{=}0.05$, and $k{=}10$. At each training step, visual prompt tokens are replaced with padding tokens with probability $p_{\text{mask}}$, and no loss is computed on masked positions. This ensures early training focuses on learning image generation from $c$ alone, while later training gradually introduces visual prompts as both targets and conditions.

\paragraph{For continuous token models.}
Since flow matching and cross-entropy losses operate at different scales, we keep visual prompts unmasked but modulate the cross-entropy loss coefficient:
\begin{equation}
\lambda_{\text{CE}}(s) = \lambda_0 \cdot (1 - \sigma(s)) + \lambda_1 \cdot \sigma(s)
\label{eq:ce_schedule}
\end{equation}
with $\lambda_0{=}0.03$, $\lambda_1{=}0.1$, and flow matching loss coefficient fixed at 1. This prevents the cross-entropy gradient from overwhelming the diffusion objective in early training. Both schedules are ablated in Appendix~\ref{app:schedule}.

\subsection{Internal vs.\ External Architectures}
\label{sec:arch}

Current image generation models can be broadly categorized into two architectural paradigms (Figure~\ref{fig:main}). \textbf{External} architectures, such as X-Omni~\cite{xomni} and BLIP3o-Next~\cite{blip3o}, use a two-stage pipeline where an autoregressive model first generates semantic tokens, which are then fed as conditioning to a separate diffusion decoder. The AR and DiT~\cite{dit} share no parameters; the DiT receives the semantic features as cross-attention conditioning. \textbf{Internal} architectures process all modalities within a single model. Transfusion~\cite{transfusion} applies causal attention for text tokens and bidirectional attention for image patches, enabling cross-modal interaction where the image generation process can directly access the conditioning information. BAGEL~\cite{bagel} further extends this with a Mixture-of-Transformers (MOT) design that uses parameter-disjoint experts within shared attention. Notably, external architectures have already begun incorporating visual semantic representations (X-Omni and BLIP3o-Next both use SigLIP 2 tokens), while internal architectures have not yet explored this direction. We therefore study how to integrate \method{} into both paradigms:

\paragraph{Internal\textsubscript{\scriptsize +\method{}}.}
Following the MOT design, the AR model handles text and SigLIP 2 tokens (cross-entropy loss), while the DiT model handles image tokens (flow matching loss). At each layer, both models use independent Q/K/V/O projections but compute attention over the \textit{concatenated} key-value sequence:
\begin{equation}
\mathbf{A} = \text{Softmax}\!\left(\frac{\mathbf{Q}_i \cdot [\mathbf{K}_1; \mathbf{K}_2]^\top}{\sqrt{d}}\right) \cdot [\mathbf{V}_1; \mathbf{V}_2], \quad i \in \{1, 2\}
\label{eq:mot_attn}
\end{equation}
where $\mathbf{Q}_i, \mathbf{K}_i, \mathbf{V}_i$ are the query, key, and value projections of model $i$. This enables the DiT model to directly attend to text, reference images, and SigLIP 2 features simultaneously.
The attention mask is causal over the AR tokens, so they never attend to the noisy image tokens. Their keys/values are therefore invariant across denoising steps and can be cached (Algorithm~\ref{alg:internal}, Appendix~\ref{app:inference}).

\paragraph{External\textsubscript{\scriptsize +\method{}}.}
The autoregressive model generates SigLIP 2 tokens from the conditioning signal, then a separate DiT model uses the SigLIP hidden features as cross-attention conditioning to generate image tokens via flow matching. External\textsubscript{\scriptsize +\method{}} is a controlled reproduction of the external paradigm of X-Omni and BLIP3o-Next with matched parameters, data, and initialization, and performs on par with these systems (Section~\ref{sec:t2i}).

In our experiments, both architectures have identical total parameter counts (4.16B for text-to-image, 4.57B for editing with reference image encoder), enabling fair comparison.

\section{Related Work}
\label{sec:related}

\paragraph{Multimodal generation.}
Recent image generation models span diverse architectures. Autoregressive models~\cite{cogview,parti,llamagen,emu3} generate images via next-token prediction over discrete tokens~\cite{vqvae,vqgan}, recent variants~\cite{var,mar,infinity} explore scalable alternatives. Diffusion and flow-based models~\cite{ddpm,flowmatching,sd3} and their transformer variants~\cite{dit,sit,imagen,pixartsigma} achieve high quality through denoising. Unified multimodal models~\cite{showo,showo2,transfusion,januspro,bagel} combine understanding and generation in a single architecture. Our work builds upon these frameworks and introduces visual prompts as a general enhancement.

\paragraph{Visual semantic tokens as intermediate representations.}
Vision-language encoders~\cite{clip,siglip2,dinov2} produce semantic features widely used as conditioning signals~\cite{dalle2,repa}. X-Omni~\cite{xomni} and BLIP3o-Next~\cite{blip3o} generate SigLIP 2 tokens to condition external diffusion models. However, these works treat visual tokens as a fixed architectural choice rather than systematically studying their impact. In contrast, we integrate visual prompts into internal architectures, investigate them across diverse frameworks and tasks, and compare internal and external integration under matched settings.

\paragraph{Image editing.}
InstructPix2Pix~\cite{instructpix2pix} pioneered instruction-based editing using classifier-free guidance over both text and image conditions. Subsequent works have improved editing through curated data and AR formulations~\cite{magicbrush,seedxedit,editar}, attention and inversion techniques~\cite{prompt2prompt,nulltext,pnp}, and unified frameworks~\cite{unicedit,pico,smartedit}. Under matched settings, our work shows that internal models that jointly attend to reference images, instructions, and semantic features can preserve details during editing, while external models whose decoder is conditioned only on semantic features preserve them poorly, and giving the external decoder direct reference access largely closes this gap (Section~\ref{sec:analysis}).

\section{Experiments}
\label{sec:exp}

We evaluate \method{} across three progressively complex tasks: class-to-image generation (\S\ref{sec:c2i}), text-to-image generation (\S\ref{sec:t2i}), and a controlled analysis of internal vs.\ external architectures for editing (\S\ref{sec:analysis}).

\subsection{Class-to-Image Generation}
\label{sec:c2i}

\begin{table}[t]
\centering
\small
\begin{minipage}[t]{0.52\textwidth}
\centering
\caption{\textbf{C2I experimental setup.}}
\label{tab:c2i_setup}
\setlength{\tabcolsep}{4pt}
\begin{tabular}{lcc}
\toprule
Framework & LlamaGen & Transfusion \\
\midrule
Initialization & --- & Qwen2.5-0.5B \\
Resolution & $512 \times 512$ & $256 \times 256$ \\
Image tokens & 1,024 & 256 \\
SigLIP 2 tokens & 196 & 121 \\
Hidden size & 1,280 & 896 \\
Model\textsubscript{\scriptsize +\method{}} params & 1.07B\textsubscript{\scriptsize \textcolor{red}{+0.04B}} & 0.37B\textsubscript{\scriptsize \textcolor{red}{+0.02B}} \\
\bottomrule
\end{tabular}
\end{minipage}%
\hfill
\begin{minipage}[t]{0.44\textwidth}
\centering
\caption{\textbf{C2I on ImageNet.} Without CFG.}
\label{tab:c2i}
\begin{tabular}{lcc}
\toprule
Method & Ep/Step & FID$\downarrow$ \\
\midrule
\multicolumn{3}{l}{\textit{Discrete ($512 \times 512$)}} \\
LlamaGen & 100 & 24.56 \\
\rowcolor{tabhighlight} \textbf{LlamaGen\textsubscript{\scriptsize +\method{}} (ours)} & 100 & \textbf{8.69} \\
\midrule
\multicolumn{3}{l}{\textit{Continuous ($256 \times 256$)}} \\
Transfusion & 266k & 15.18 \\
\rowcolor{tabhighlight} \textbf{Transfusion\textsubscript{\scriptsize +\method{}} (ours)} & 266k & \textbf{11.43} \\
\bottomrule
\end{tabular}
\end{minipage}
\end{table}

\paragraph{Setup.}
We train on ImageNet-1K~\cite{imagenet} with two frameworks (Table~\ref{tab:c2i_setup}) to validate the effectiveness of \method{} under both discrete and continuous schemes. \method{} adds only a randomly initialized SigLIP 2 embedding layer ($<4\%$ parameter increase). All results are evaluated on the ImageNet validation set without classifier-free guidance (CFG$=$1.0).

\begin{itemize}[nosep,leftmargin=*]
\item \textbf{Discrete.} We adopt LlamaGen~\cite{llamagen} with the Emu3~\cite{emu3} tokenizer to generate $512{\times}512$ images (white-padded) as 1,024 tokens. LlamaGen\textsubscript{\scriptsize +\method{}} prepends 196 SigLIP 2 tokens (from the $224{\times}224$ white-padded target image), increasing per-sample length from 1,025 to 1,221 tokens. Trained from scratch for 100 epochs with batch size 128 for both LlamaGen and LlamaGen\textsubscript{\scriptsize +\method{}}, where the latter consumes 19.1\% more tokens per epoch.
\item \textbf{Continuous.} We adopt Transfusion~\cite{transfusion} to generate $256{\times}256$ images (center-cropped) as 256 VAE tokens, with fixed-length sequence packing~\cite{bagel} at 10,240 tokens per step. Transfusion\textsubscript{\scriptsize +\method{}} prepends 121 SigLIP 2 tokens (from the $176{\times}176$ center-cropped target image) but sees approximately 32\% fewer images per step due to longer per-sample sequences. The model loads non-embedding weights from Qwen2.5-0.5B and randomly initializes 1,000 class token embeddings, resulting in 0.37B total parameters. Trains 130k shared steps without \method{}, then forks into Transfusion and Transfusion\textsubscript{\scriptsize +\method{}} branches for 136k additional steps.
\end{itemize}

\paragraph{Results.} \method{} consistently improves final generation quality and convergence speed across LlamaGen and Transfusion, validating its effectiveness under both discrete and continuous token paradigms.

\begin{figure}[t]
\centering
\includegraphics[width=\linewidth]{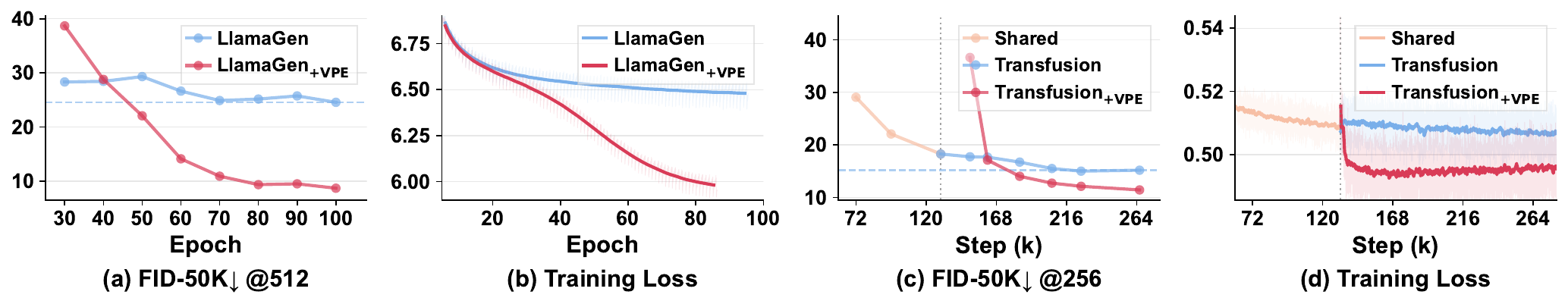}
\caption{\textbf{Convergence on ImageNet.} \textbf{(a)} LlamaGen FID-50K ($512{\times}512$): \method{} starts higher due to progressive masking ($p_{\text{mask}}{=}0.95$) but surpasses the baseline from epoch 50, with the gap widening to 15.87 by epoch 100. \textbf{(b)} LlamaGen training loss converges to a lower value with \method{}. \textbf{(c)} Transfusion FID-50K ($256{\times}256$): both branches fork from the shared checkpoint (dashed line). The initial high FID of \method{} at 150k reflects the train-inference gap on self-generated visual prompts, which resolves quickly. \textbf{(d)} Transfusion flow loss curves for the shared base and both branches.}
\label{fig:c2i_conv}
\end{figure}

\begin{itemize}[nosep,leftmargin=*]

\item \textbf{Quality ceiling (Table~\ref{tab:c2i}, Figure~\ref{fig:c2i_conv}).}
At comparable training budgets, \method{} substantially raises the quality ceiling. LlamaGen\textsubscript{\scriptsize +\method{}} achieves FID 8.69 vs.\ 24.56 at epoch 100. Transfusion\textsubscript{\scriptsize +\method{}} achieves FID 11.43 vs.\ 15.18 at step 266k. The training loss curves (Figure~\ref{fig:c2i_conv}b,d) confirm that \method{} converges to a lower loss in both frameworks.

\item \textbf{Convergence acceleration (Figure~\ref{fig:c2i_conv}).}
\method{} dramatically accelerates convergence. As shown in Figure~\ref{fig:c2i_conv}a, LlamaGen\textsubscript{\scriptsize +\method{}} at epoch 50 already surpasses LlamaGen at epoch 100 (FID 22.07 vs.\ 24.56). Since \method{} adds 19.1\% more tokens per epoch, LlamaGen\textsubscript{\scriptsize +\method{}} at epoch 50 consumes approximately 60\% of LlamaGen's total token budget at epoch 100. For Transfusion, where both branches consume the same tokens per step, Transfusion\textsubscript{\scriptsize +\method{}} at step 184k surpasses Transfusion at step 266k (FID 14.02 vs.\ 15.18, Figure~\ref{fig:c2i_conv}c). This illustrates that \method{} effectively reduces the difficulty of directly modeling images from text.

\item \textbf{Early-stage behavior (Figure~\ref{fig:c2i_conv}).}
LlamaGen\textsubscript{\scriptsize +\method{}}'s early-epoch FID increase (epochs 30--40) is expected: progressive masking ($p_{\text{mask}}{=}0.95$) limits the visual prompt signal initially. Once the model learns to leverage visual prompts (epoch 50+), improvement is decisive and monotonic. For Transfusion\textsubscript{\scriptsize +\method{}}, the initial high FID at step 150k reflects the train-inference gap: training uses ground-truth SigLIP 2 tokens while inference relies on self-generated ones. This resolves quickly as the model learns to produce reliable visual prompts.

\item \textbf{Further analysis.} Ablations of the progressive schedule, a comparison between ground-truth and predicted visual prompts, and training cost are given in Appendices~\ref{app:schedule}, \ref{app:gt_vp}, and~\ref{app:efficiency}.

\end{itemize}

\subsection{Text-to-Image Generation}
\label{sec:t2i}

\begin{table}[t]
\caption{\textbf{Text rendering on TextAtlas benchmarks.} Metrics include CLIP Score (CS) $\uparrow$, OCR Accuracy (Acc.) $\uparrow$, F1 Score $\uparrow$, and Character Error Rate (CER) $\downarrow$. All results use CFG$=$5.5. Best open-source values in \textbf{bold}. \colorbox{closedmodel}{Gray}: closed-source models. $*$: continued training on 6.35M text rendering data.}
\label{tab:textatlas}
\centering
\small
\setlength{\tabcolsep}{2.5pt}
\begin{tabular}{lcccccccccccccc}
\toprule
& \multicolumn{4}{c}{TextScenesHQ} & & \multicolumn{4}{c}{TextVisionBlend} & & \multicolumn{4}{c}{StyleTextSynth} \\
\cmidrule(lr){2-5} \cmidrule(lr){7-10} \cmidrule(lr){12-15}
Method & CS$\uparrow$ & Acc$\uparrow$ & F1$\uparrow$ & CER$\downarrow$ && CS$\uparrow$ & Acc$\uparrow$ & F1$\uparrow$ & CER$\downarrow$ && CS$\uparrow$ & Acc$\uparrow$ & F1$\uparrow$ & CER$\downarrow$ \\
\midrule
\rowcolor{closedmodel} Grok~3 & 0.32 & 35.07 & 37.94 & 0.57 && 0.17 & 41.54 & 44.22 & 0.57 && 0.29 & 15.82 & 21.40 & 0.73 \\
\rowcolor{closedmodel} DALL-E~3~\cite{dalle3} & 0.34 & 69.26 & 51.63 & 0.67 && 0.19 & 8.38 & 7.94 & 0.93 && 0.29 & 30.58 & 38.25 & 0.78 \\
\midrule
AnyText~\cite{anytext} & 0.22 & 0.42 & 0.81 & 0.95 && -- & -- & -- & -- && 0.25 & 0.35 & 0.66 & 0.98 \\
TextDiffuser-2~\cite{textdiffuser2} & 0.23 & 0.66 & 1.25 & 0.96 && -- & -- & -- & -- && 0.25 & 0.76 & 1.46 & 0.99 \\
PixArt-$\Sigma$~\cite{pixartsigma} & 0.23 & 0.34 & 0.53 & 0.91 && 0.19 & 2.40 & 1.57 & 0.83 && 0.28 & 0.42 & 0.62 & 0.90 \\
Infinity-2B~\cite{infinity} & 0.23 & 1.06 & 1.74 & 0.88 && \textbf{0.20} & 2.98 & 3.44 & 0.83 && 0.27 & 0.80 & 1.42 & 0.93 \\
SD3.5 Large~\cite{sd3} & 0.24 & \textbf{19.03} & 24.45 & \textbf{0.73} && 0.18 & 14.55 & 16.25 & 0.88 && 0.28 & 27.21 & 33.86 & 0.73 \\
\midrule
Show-o2$^*$ (1.5B)~\cite{showo2} & \textbf{0.29} & 2.62 & 4.36 & 0.88 && 0.16 & 76.45 & 77.40 & 0.22 && \textbf{0.29} & 21.83 & 27.72 & \textbf{0.59} \\
\rowcolor{tabhighlight} \textbf{Show-o2\textsubscript{\scriptsize +\method{}} (ours, 1.5B)} & 0.26 & 16.29 & \textbf{26.07} & 0.77 && 0.16 & \textbf{80.07} & \textbf{82.55} & \textbf{0.20} && 0.26 & \textbf{32.91} & \textbf{43.78} & 0.60 \\
\bottomrule
\end{tabular}
\end{table}

\begin{figure}[t]
\centering
\includegraphics[width=0.82\linewidth]{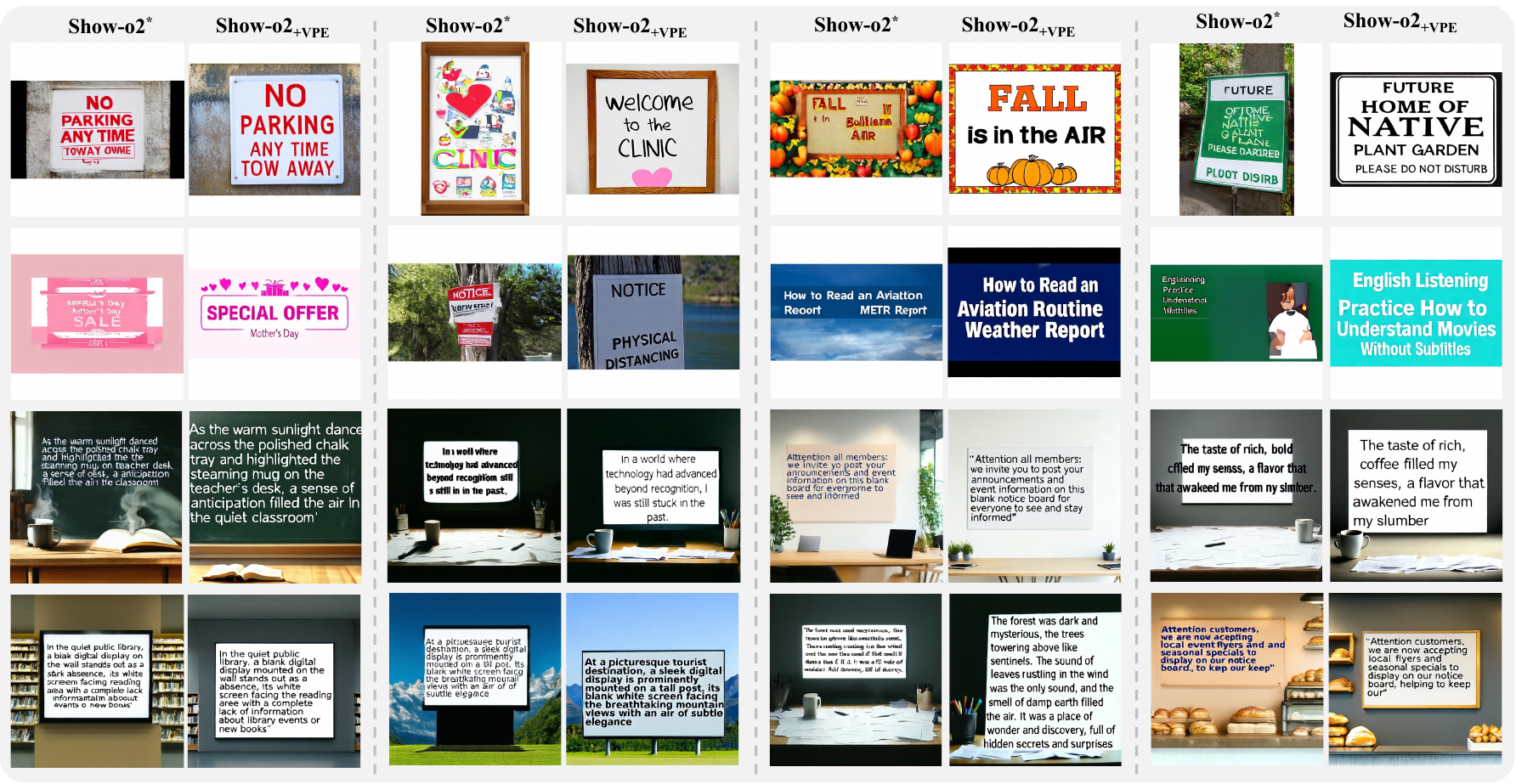} 
\caption{\textbf{Text rendering comparison.} Show-o2\textsubscript{\scriptsize +\method{}} generates more accurate text compared to Show-o2$^*$, consistent with the quantitative improvements in Table~\ref{tab:textatlas}.}
\label{fig:text_rendering}
\end{figure}

\begin{table}[t]
\caption{\textbf{GenEval benchmark.} External\textsubscript{\scriptsize +\method{}} at 4.16B surpasses many larger models, demonstrating the effectiveness of \method{} for T2I generation quality.}
\label{tab:geneval}
\centering
\small
\begin{tabular}{lcccccccc}
\toprule
Method & \#Params & Single & Two & Count & Color & Pos. & C.Attr. & Overall$\uparrow$ \\
\midrule
ILLUME~\cite{illume} & 7B & 0.99 & 0.86 & 0.45 & 0.71 & 0.39 & 0.28 & 0.61 \\
Transfusion~\cite{transfusion} & 7B & -- & -- & -- & -- & -- & -- & 0.63 \\
D-DiT~\cite{ddit} & 2B & 0.97 & 0.80 & 0.54 & 0.76 & 0.32 & 0.50 & 0.65 \\
Emu3~\cite{emu3} & 8B & -- & -- & -- & -- & -- & -- & 0.66 \\
SD3-Medium~\cite{sd3} & -- & 0.99 & 0.94 & 0.72 & 0.89 & 0.33 & 0.60 & 0.74 \\
Show-o2~\cite{showo2} & 7B & \textbf{1.00} & 0.87 & 0.58 & \textbf{0.92} & 0.52 & 0.62 & 0.76 \\
BLIP3o-NEXT-SFT~\cite{blip3o} & 3B & 0.98 & 0.88 & 0.52 & 0.86 & 0.74 & 0.61 & 0.77 \\
MetaQuery-XL~\cite{metaqueryxl} & 7B & -- & -- & -- & -- & -- & -- & 0.80 \\
Janus-Pro~\cite{januspro} & 7B & 0.99 & 0.89 & 0.59 & 0.90 & \textbf{0.79} & 0.66 & 0.80 \\
X-Omni~\cite{xomni} & 7B & 0.98 & \textbf{0.95} & \textbf{0.75} & 0.91 & 0.71 & \textbf{0.68} & \textbf{0.83} \\
\midrule
\rowcolor{tabhighlight} \textbf{Internal\textsubscript{\scriptsize +\method{}} (ours)} & 4.16B & 0.98 & 0.91 & 0.68 & 0.86 & 0.77 & 0.55 & 0.79 \\
\rowcolor{tabhighlight} \textbf{External\textsubscript{\scriptsize +\method{}} (ours)} & 4.16B & 0.98 & 0.90 & 0.69 & 0.86 & 0.76 & \textbf{0.68} & 0.81 \\
\bottomrule
\end{tabular}
\end{table}

\paragraph{Text rendering setup.}
\label{sec:textrender}
We fine-tune the pre-trained Show-o2-1.5B model~\cite{showo2} on 6.35M English text rendering images collected from TextAtlas~\cite{textatlas} and AnyWord-3M~\cite{anytext}, excluding Chinese text images and academic paper images whose text becomes illegible after white-padding preprocessing. Since the official Show-o2-1.5B was not trained on dense text data~\cite{showo2}, we also continue-train the baseline on the same 6.35M data (denoted Show-o2$^*$) to ensure improvements stem from \method{} rather than additional data. Show-o2\textsubscript{\scriptsize +\method{}} uses batch size 8 with max sequence length 1,856, while Show-o2$^*$ uses batch size 12 with max sequence length 1,280, resulting in closely matched per-step token throughput ($<3.5\%$ difference). Both models are trained for 3 epochs on 24 H200 GPUs. At this matched token throughput, the two runs have nearly identical cost (2,186 vs.\ 2,224 GPU-hours, Appendix~\ref{app:efficiency}). The \method{} model's CE loss follows Eq.~\ref{eq:ce_schedule} with total training steps $S{=}100$k.

\paragraph{Text rendering results.} Show-o2\textsubscript{\scriptsize +\method{}} achieves notable improvements over existing open-source models and Show-o2$^*$, demonstrating that visual prompts effectively reduce the difficulty of directly modeling text-to-image within internal architectures such as Show-o2.

\begin{itemize}[nosep,leftmargin=*]
\item \textbf{Comparison with open-source models (Table~\ref{tab:textatlas}).}
Show-o2\textsubscript{\scriptsize +\method{}} at 1.5B parameters outperforms all open-source baselines. It only slightly trails SD3.5 Large on TextScenesHQ, but given that SD3.5 Large has 8.1B parameters ($5.4\times$ larger), this already demonstrates strong competitiveness. On TextVisionBlend, it achieves the highest accuracy (80.07\%) and F1 (82.55) among all methods including closed-source ones.

\item \textbf{Controlled comparison with Show-o2$^*$ (Table~\ref{tab:textatlas}).}
Compared to Show-o2$^*$ trained on identical data without \method{}, Show-o2\textsubscript{\scriptsize +\method{}} improves OCR accuracy from 2.62\% to 16.29\% ($6.2\times$) on TextScenesHQ, from 76.45\% to 80.07\% on TextVisionBlend, and from 21.83\% to 32.91\% on StyleTextSynth. These consistent improvements confirm that visual prompts help the model ``plan'' text layout before rendering. By first generating SigLIP 2 tokens that encode the semantic content, the model produces more accurate character structures during the subsequent generation phase.

\item \textbf{Qualitative results (Figure~\ref{fig:text_rendering}).}
Show-o2\textsubscript{\scriptsize +\method{}} generates more accurate and readable text compared to Show-o2$^*$, consistent with the quantitative improvements.

\item \textbf{Image quality.} \method{} also improves the LAION aesthetic score (Appendix~\ref{app:t2i_extra}).
\end{itemize}

\paragraph{General T2I setup.}
Beyond text rendering, we evaluate whether \method{} improves general T2I quality. We adopt two mainstream architectures illustrated in Figure~\ref{fig:main}: Internal\textsubscript{\scriptsize +\method{}} (mixture-of-transformers) and External\textsubscript{\scriptsize +\method{}} (AR + DiT). Both models load all parameters from Show-o2-1.5B~\cite{showo2} except for the lightweight SigLIP 2 embedding layer, which is randomly initialized. Since the T2I task has no image input, we remove the VAE encoder, resulting in 4.16B parameters.
We collect 32M T2I images from the publicly available BLIP3o~\cite{blip3o} data, combined with the 6.35M text rendering data described above. Training proceeds in two stages:
\begin{itemize}[nosep,leftmargin=*]
\item \textbf{Stage-1} (text-to-SigLIP 2 alignment): trains only the AR model for 4 epochs on the 39M mixed data. The resulting AR checkpoint is shared by both architectures in Stage-2, ensuring identical text-to-SigLIP 2 alignment capability.
\item \textbf{Stage-2} (image generation): trains on the same 39M mixed data, following each framework's standard training paradigm. Internal\textsubscript{\scriptsize +\method{}} performs joint training of both the AR and DiT models. External\textsubscript{\scriptsize +\method{}} freezes the pretrained AR model and only trains the DiT.
\end{itemize}
This setup also serves as the foundation for the editing experiments in Section~\ref{sec:analysis}.

\paragraph{General T2I results.}
Table~\ref{tab:geneval} shows GenEval~\cite{geneval} results. At only 4.16B parameters, External\textsubscript{\scriptsize +\method{}} surpasses most compared 7B+ models on overall score, performs on par with X-Omni (0.83, 7B, with RL post-training), and outperforms BLIP3o-NEXT-SFT (0.77, 3B), while Internal\textsubscript{\scriptsize +\method{}} remains competitive with 7B-scale models. This also confirms that our controlled External\textsubscript{\scriptsize +\method{}} is a faithful, competitive instance of the external paradigm. This illustrates that \method{} reduces the modeling difficulty and improves generation quality, enabling smaller models to exceed much larger ones. On GenAI-Bench~\cite{genaibench}, DSG-1k~\cite{dsg}, and GenEval with 3 seeds (Table~\ref{tab:t2i_extra}), both models match T2I-R1~\cite{t2ir1} and perform on par with each other.

\subsection{Analysis: Internal vs.\ External architecture}
\label{sec:analysis}

\begin{figure}[t]
\centering
\includegraphics[width=0.82\linewidth]{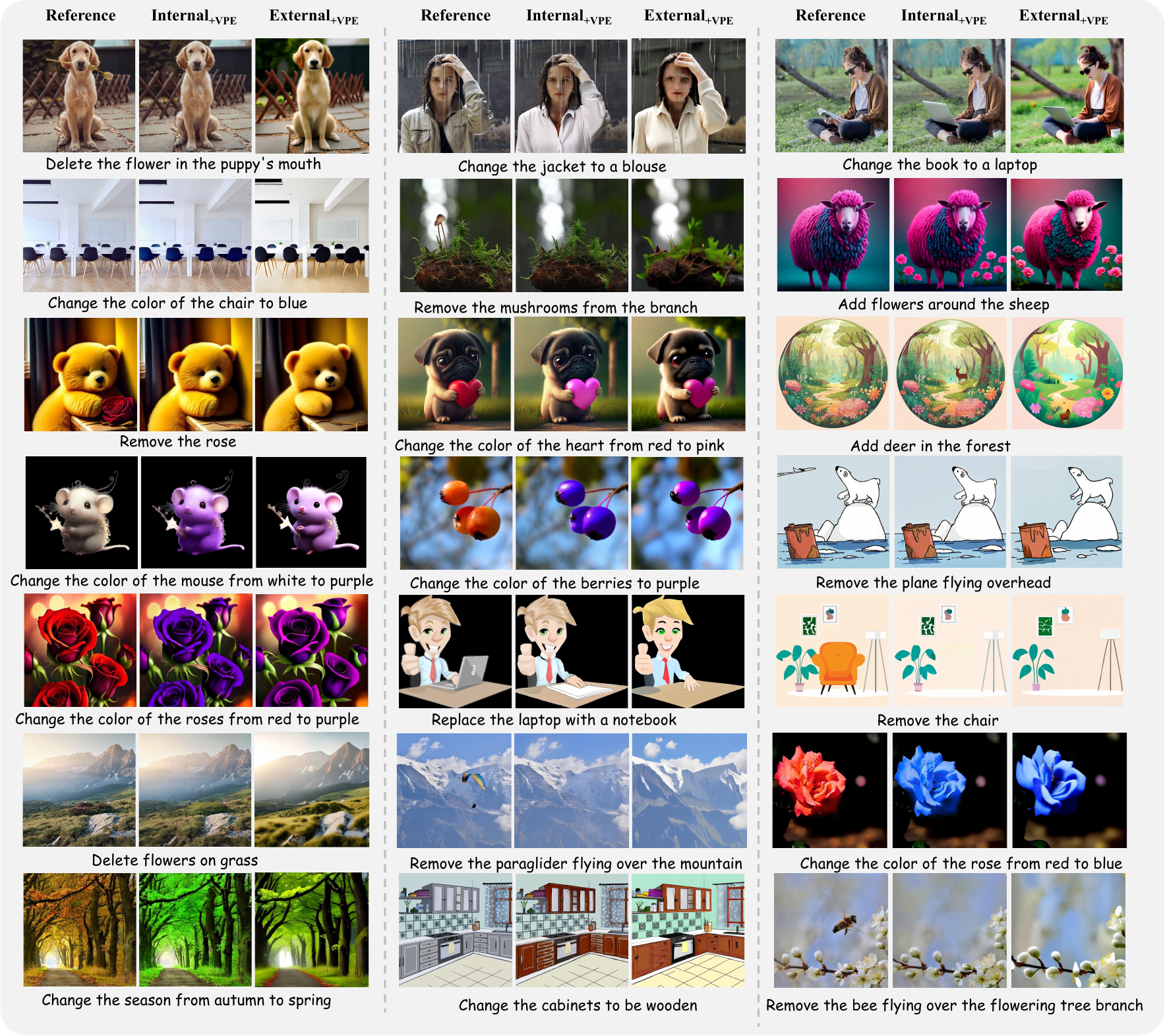} 
\caption{\textbf{Editing comparison.} Each triplet shows Reference / Internal\textsubscript{\scriptsize +\method{}} / External\textsubscript{\scriptsize +\method{}}. Internal\textsubscript{\scriptsize +\method{}} preserves unedited regions while External\textsubscript{\scriptsize +\method{}} regenerates inconsistent backgrounds.}
\label{fig:editing_main}
\end{figure}

\begin{table}[t]
\caption{\textbf{Internal vs.\ External architecture comparison.} Both architectures achieve comparable T2I quality ((a), Table~\ref{tab:geneval}, and Table~\ref{tab:t2i_extra}), but Internal\textsubscript{\scriptsize +\method{}} dramatically outperforms External\textsubscript{\scriptsize +\method{}} in editing preservation (b). In (b), the ``+'' rows are ablations of External\textsubscript{\scriptsize +\method{}}. They start from the same Stage-2 checkpoint, fully unfrozen, with the same data and 70k steps. ``+ ref access'' also feeds the reference image, text, and SigLIP 2 features to the DiT decoder. BLIP3o-NEXT uses its released editing weights.}
\label{tab:editing}
\centering
\small

\textit{(a) TextAtlas (text rendering, CFG$=$5.5)}\\[2pt]
\setlength{\tabcolsep}{3pt}
\begin{tabular}{lcccccccccccc}
\toprule
& \multicolumn{4}{c}{TextScenesHQ} & \multicolumn{4}{c}{TextVisionBlend} & \multicolumn{4}{c}{StyleTextSynth} \\
\cmidrule(lr){2-5} \cmidrule(lr){6-9} \cmidrule(lr){10-13}
Method & CS & Acc$\uparrow$ & F1$\uparrow$ & CER$\downarrow$ & CS & Acc$\uparrow$ & F1$\uparrow$ & CER$\downarrow$ & CS & Acc$\uparrow$ & F1$\uparrow$ & CER$\downarrow$ \\
\midrule
Internal\textsubscript{\scriptsize +\method{}} & 0.27 & \textbf{22.70} & \textbf{34.37} & \textbf{0.72} & 0.16 & 77.59 & \textbf{80.25} & 0.23 & 0.27 & \textbf{38.63} & \textbf{49.62} & 0.54 \\
External\textsubscript{\scriptsize +\method{}} & 0.27 & 19.07 & 29.69 & 0.75 & 0.16 & 77.59 & 80.15 & 0.23 & 0.27 & 38.24 & 49.06 & 0.54 \\
\bottomrule
\end{tabular}

\vspace{6pt}
\textit{(b) PIE-Bench (Editing, Image CFG$=$1.0, Text CFG$=$2.5)}\\[2pt]
\setlength{\tabcolsep}{3.5pt}
\begin{tabular}{lccccccc}
\toprule
Method & \makecell{Struct.\\Dist.$\downarrow$} & PSNR$\uparrow$ & LPIPS$\downarrow$ & MSE$\downarrow$ & SSIM$\uparrow$ & \makecell{CLIP\\Whole$\uparrow$} & \makecell{CLIP\\Edit$\uparrow$} \\
\midrule
Internal\textsubscript{\scriptsize +\method{}} & \textbf{24.60} & \textbf{26.76} & \textbf{58.61} & \textbf{46.76} & \textbf{86.66} & 23.00 & 20.50 \\
\midrule
BLIP3o-NEXT~\cite{blip3o} & 98.87 & 15.94 & 198.84 & 333.50 & 62.92 & 25.09 & \textbf{22.28} \\
External\textsubscript{\scriptsize +\method{}} & 61.66 & 19.92 & 158.09 & 150.88 & 70.02 & 23.09 & 20.69 \\
\quad + unfrozen & 62.32 & 18.56 & 168.66 & 203.68 & 68.90 & \textbf{25.34} & 21.89 \\
\quad + unfrozen + ref access & 26.67 & 25.76 & 68.66 & 53.86 & 81.98 & 23.56 & 20.63 \\
\bottomrule
\end{tabular}
\end{table}

\paragraph{Setup.}
\label{sec:edit}
Both Internal\textsubscript{\scriptsize +\method{}} and External\textsubscript{\scriptsize +\method{}} from Section~\ref{sec:t2i} achieve comparable T2I quality (0.79 and 0.81 on GenEval), providing a relatively fair baseline for comparing internal and external paradigms on editing. Full training details are provided in Appendix~\ref{app:training}. Both models add a reference image encoder (initialized from Show-o2-1.5B~\cite{showo2} weights), bringing total parameters to 4.57B with the VAE encoder included. Starting from the Stage-2 T2I checkpoints, we fine-tune on 5.4M editing pairs from NHR-Edit~\cite{nhr} ($\times$3 upsampling, 2.16M), UnicEdit~\cite{unicedit} (2.03M), ShareGPT4o~\cite{sharegpt4o} ($\times$5, 233k), and Pico-Banana~\cite{pico} ($\times$4, 1.03M), with batch size 5 per GPU on 24 H200 GPUs for approximately 1.5 epochs (70k steps). We adopt InstructPix2Pix-style~\cite{instructpix2pix} conditional dropout for text and reference image conditions, where both models drop text and SigLIP 2 tokens \textit{jointly}, and the no-text, no-reference, and null cases are mutually exclusive (5\% each) (details in Appendix~\ref{app:editing_cfg}). Inference uses classifier-free guidance with text CFG 2.5 and image CFG 1.0. We evaluate on PIE-Bench~\cite{pnp}.

\paragraph{Results.}
Table~\ref{tab:editing} shows that while both architectures achieve comparable T2I quality and editing \textit{responsiveness} (similar CLIP similarity scores), Internal\textsubscript{\scriptsize +\method{}} dramatically outperforms External\textsubscript{\scriptsize +\method{}} in detail \textit{preservation}, with $2.5\times$ better Structure Distance (24.60 vs.\ 61.66), $+6.84$ dB PSNR (26.76 vs.\ 19.92), $2.7\times$ better LPIPS (58.61 vs.\ 158.09), and $+16.6$ SSIM (86.66 vs.\ 70.02). This gap arises despite identical parameter counts, training data, and training duration. The ablations in Table~\ref{tab:editing}b identify its source: unfreezing the external model does not close the gap (PSNR 18.56), while additionally routing the reference image, text, and SigLIP 2 features into the DiT decoder reduces it to 1.0\,dB (PSNR 25.76 vs.\ 26.76). The gap therefore stems from the semantic-only DiT conditioning adopted by existing external systems such as X-Omni and BLIP3o-Next. On I2EBench~\cite{i2ebench} (Appendix~\ref{app:edit_extra}), Internal\textsubscript{\scriptsize +\method{}} wins all low-level dimensions and is comparable on high-level ones. Figure~\ref{fig:editing_main} shows representative examples: Internal\textsubscript{\scriptsize +\method{}} faithfully preserves unedited regions while External\textsubscript{\scriptsize +\method{}} regenerates the entire scene.

\label{sec:theory}
We provide an explanation for this gap from an information perspective, for an external decoder conditioned only on semantic features. Let $\mathbf{x}$ denote the reference image, $t$ the edit instruction, and $\mathcal{P} \subseteq \mathbf{x}$ the region that should remain unchanged. An ideal edit modifies only the attributes specified by $t$ while keeping $\mathcal{P}$ intact.

In the external pipeline, the AR model takes the reference image (via VAE encoding) and the edit instruction, producing $\mathbf{s}' = f_{\text{AR}}(\mathbf{x}, t)$, the SigLIP 2 representation of the target image after editing. Since SigLIP 2 is a semantic encoder that discards fine-grained spatial details~\cite{siglip2}, $\mathbf{s}'$ inevitably loses the pixel-level information (textures, edges, background details) that must be preserved. The DiT decoder generates $\hat{\mathbf{x}} = g_{\text{DiT}}(\mathbf{s}')$, and by the data processing inequality:
\begin{equation}
I(\hat{\mathbf{x}}; \mathcal{P}) \leq I(\mathbf{s}'; \mathcal{P}) < H(\mathcal{P})
\label{eq:bottleneck}
\end{equation}
The first inequality holds because $\hat{\mathbf{x}}$ is a function of $\mathbf{s}'$. The strict inequality holds because $\mathbf{s}'$ is a SigLIP 2 representation that has irreversibly discarded the fine-grained details of $\mathcal{P}$.

In contrast, the internal design allows the DiT to attend jointly to the full sequence $[\mathbf{x}, t, \mathbf{s}']$ at every layer through shared attention (Eq.~\ref{eq:mot_attn}):
\begin{equation}
I(\hat{\mathbf{x}}; \mathcal{P}) \leq I(\mathbf{x}, t, \mathbf{s}'; \mathcal{P}) = H(\mathcal{P})
\label{eq:internal}
\end{equation}
since $\mathcal{P} \subseteq \mathbf{x}$. The model can cross-reference $t$ (what to change) with $\mathbf{x}$ (original details) at every decoding step, using $\mathbf{s}'$ purely for semantic guidance. No information bottleneck exists because the decoder has direct access to $\mathbf{x}$ throughout the generation process. An external decoder given direct reference access likewise escapes the first bound, matching the ablation above.
\section{Conclusion}
\label{sec:conclusion}

We have presented \methodfull{} (\method{}), which integrates visual semantic planning into unified architectures, together with the progressive training schedule this requires.
Our systematic study demonstrates that \method{} accelerates convergence and raises quality ceilings across discrete and continuous token types, significantly enhances text rendering, and through internal architecture integration enables effective image editing with strong detail preservation.
Under the evaluated configurations, internal architectures with joint self-attention preserve editing details far better than external ones with semantic-only DiT conditioning: internal joint attention provides reference access natively, while external pipelines must inject it explicitly.

\begin{ack}
This research is supported by National Research Foundation, Singapore, NRF-NRFI10-2024-0004. The authors declare no competing interests.
\end{ack}


\bibliographystyle{unsrt}
\bibliography{references}

@article{cot,
  title={Chain-of-thought prompting elicits reasoning in large language models},
  author={Wei, Jason and Wang, Xuezhi and Schuurmans, Dale and Bosma, Maarten and Xia, Fei and Chi, Ed and Le, Quoc V and Zhou, Denny and others},
  journal={Advances in neural information processing systems},
  volume={35},
  pages={24824--24837},
  year={2022}
}

@article{siglip2,
  title={Siglip 2: Multilingual vision-language encoders with improved semantic understanding, localization, and dense features},
  author={Tschannen, Michael and Gritsenko, Alexey and Wang, Xiao and Naeem, Muhammad Ferjad and Alabdulmohsin, Ibrahim and Parthasarathy, Nikhil and Evans, Talfan and Beyer, Lucas and Xia, Ye and Mustafa, Basil and others},
  journal={arXiv preprint arXiv:2502.14786},
  year={2025}
}

@article{llamagen,
  title={Autoregressive model beats diffusion: Llama for scalable image generation},
  author={Sun, Peize and Jiang, Yi and Chen, Shoufa and Zhang, Shilong and Peng, Bingyue and Luo, Ping and Yuan, Zehuan},
  journal={arXiv preprint arXiv:2406.06525},
  year={2024}
}

@article{emu3,
  title={Emu3: Next-token prediction is all you need},
  author={Wang, Xinlong and Zhang, Xiaosong and Luo, Zhengxiong and Sun, Quan and Cui, Yufeng and Wang, Jinsheng and Zhang, Fan and Wang, Yueze and Li, Zhen and Yu, Qiying and others},
  journal={arXiv preprint arXiv:2409.18869},
  year={2024}
}

@article{transfusion,
  title={Transfusion: Predict the next token and diffuse images with one multi-modal model},
  author={Zhou, Chunting and Yu, Lili and Babu, Arun and Tirumala, Kushal and Yasunaga, Michihiro and Shamis, Leonid and Kahn, Jacob and Ma, Xuezhe and Zettlemoyer, Luke and Levy, Omer},
  journal={arXiv preprint arXiv:2408.11039},
  year={2024}
}

@article{bagel,
  title={Emerging properties in unified multimodal pretraining},
  author={Deng, Chaorui and Zhu, Deyao and Li, Kunchang and Gou, Chenhui and Li, Feng and Wang, Zeyu and Zhong, Shu and Yu, Weihao and Nie, Xiaonan and Song, Ziang and others},
  journal={arXiv preprint arXiv:2505.14683},
  year={2025}
}

@article{showo,
  title={Show-o: One single transformer to unify multimodal understanding and generation},
  author={Xie, Jinheng and Mao, Weijia and Bai, Zechen and Zhang, David Junhao and Wang, Weihao and Lin, Kevin Qinghong and Gu, Yuchao and Chen, Zhijie and Yang, Zhenheng and Shou, Mike Zheng},
  journal={arXiv preprint arXiv:2408.12528},
  year={2024}
}

@article{showo2,
  title={Show-o2: Improved native unified multimodal models},
  author={Xie, Jinheng and Yang, Zhenheng and Shou, Mike Zheng},
  journal={arXiv preprint arXiv:2506.15564},
  year={2025}
}

@article{xomni,
  title={X-omni: Reinforcement learning makes discrete autoregressive image generative models great again},
  author={Geng, Zigang and Wang, Yibing and Ma, Yeyao and Li, Chen and Rao, Yongming and Gu, Shuyang and Zhong, Zhao and Lu, Qinglin and Hu, Han and Zhang, Xiaosong and others},
  journal={arXiv preprint arXiv:2507.22058},
  year={2025}
}

@article{blip3o,
  title={Blip3o-next: Next frontier of native image generation},
  author={Chen, Jiuhai and Xue, Le and Xu, Zhiyang and Pan, Xichen and Yang, Shusheng and Qin, Can and Yan, An and Zhou, Honglu and Chen, Zeyuan and Huang, Lifu and others},
  journal={arXiv preprint arXiv:2510.15857},
  year={2025}
}

@inproceedings{dit,
  title={Scalable diffusion models with transformers},
  author={Peebles, William and Xie, Saining},
  booktitle={Proceedings of the IEEE/CVF international conference on computer vision},
  pages={4195--4205},
  year={2023}
}

@inproceedings{sit,
  title={Sit: Exploring flow and diffusion-based generative models with scalable interpolant transformers},
  author={Ma, Nanye and Goldstein, Mark and Albergo, Michael S and Boffi, Nicholas M and Vanden-Eijnden, Eric and Xie, Saining},
  booktitle={European Conference on Computer Vision},
  pages={23--40},
  year={2024},
  organization={Springer}
}

@article{repa,
  title={Representation alignment for generation: Training diffusion transformers is easier than you think},
  author={Yu, Sihyun and Kwak, Sangkyung and Jang, Huiwon and Jeong, Jongheon and Huang, Jonathan and Shin, Jinwoo and Xie, Saining},
  journal={arXiv preprint arXiv:2410.06940},
  year={2024}
}

@article{flowmatching,
  title={Flow matching for generative modeling},
  author={Lipman, Yaron and Chen, Ricky TQ and Ben-Hamu, Heli and Nickel, Maximilian and Le, Matt},
  journal={arXiv preprint arXiv:2210.02747},
  year={2022}
}

@inproceedings{clip,
  title={Learning transferable visual models from natural language supervision},
  author={Radford, Alec and Kim, Jong Wook and Hallacy, Chris and Ramesh, Aditya and Goh, Gabriel and Agarwal, Sandhini and Sastry, Girish and Askell, Amanda and Mishkin, Pamela and Clark, Jack and others},
  booktitle={International conference on machine learning},
  pages={8748--8763},
  year={2021},
  organization={PmLR}
}

@inproceedings{imagenet,
  title={Imagenet: A large-scale hierarchical image database},
  author={Deng, Jia and Dong, Wei and Socher, Richard and Li, Li-Jia and Li, Kai and Fei-Fei, Li},
  booktitle={2009 IEEE conference on computer vision and pattern recognition},
  pages={248--255},
  year={2009},
  organization={Ieee}
}

@inproceedings{instructpix2pix,
  title={Instructpix2pix: Learning to follow image editing instructions},
  author={Brooks, Tim and Holynski, Aleksander and Efros, Alexei A},
  booktitle={Proceedings of the IEEE/CVF conference on computer vision and pattern recognition},
  pages={18392--18402},
  year={2023}
}

@article{magicbrush,
  title={Magicbrush: A manually annotated dataset for instruction-guided image editing},
  author={Zhang, Kai and Mo, Lingbo and Chen, Wenhu and Sun, Huan and Su, Yu},
  journal={Advances in Neural Information Processing Systems},
  volume={36},
  pages={31428--31449},
  year={2023}
}

@article{pnp,
  title={Direct inversion: Boosting diffusion-based editing with 3 lines of code},
  author={Ju, Xuan and Zeng, Ailing and Bian, Yuxuan and Liu, Shaoteng and Xu, Qiang},
  journal={arXiv preprint arXiv:2310.01506},
  year={2023}
}

@inproceedings{editar,
  title={Editar: Unified conditional generation with autoregressive models},
  author={Mu, Jiteng and Vasconcelos, Nuno and Wang, Xiaolong},
  booktitle={Proceedings of the Computer Vision and Pattern Recognition Conference},
  pages={7899--7909},
  year={2025}
}

@article{seedxedit,
  title={Seed-x: Multimodal models with unified multi-granularity comprehension and generation},
  author={Ge, Yuying and Zhao, Sijie and Zhu, Jinguo and Ge, Yixiao and Yi, Kun and Song, Lin and Li, Chen and Ding, Xiaohan and Shan, Ying},
  journal={arXiv preprint arXiv:2404.14396},
  year={2024}
}

@article{dalle3,
  title={Improving image generation with better captions},
  author={Betker, James and Goh, Gabriel and Jing, Li and Brooks, Tim and Wang, Jianfeng and Li, Linjie and Ouyang, Long and Zhuang, Juntang and Lee, Joyce and Guo, Yufei and others},
  journal={Computer Science. https://cdn. openai. com/papers/dall-e-3. pdf},
  volume={2},
  number={3},
  pages={8},
  year={2023}
}

@inproceedings{sd3,
  title={Scaling rectified flow transformers for high-resolution image synthesis},
  author={Esser, Patrick and Kulal, Sumith and Blattmann, Andreas and Entezari, Rahim and M{\"u}ller, Jonas and Saini, Harry and Levi, Yam and Lorenz, Dominik and Sauer, Axel and Boesel, Frederic and others},
  booktitle={Forty-first international conference on machine learning},
  year={2024}
}

@article{textatlas,
  title={Textatlas5m: A large-scale dataset for dense text image generation},
  author={Wang, Alex Jinpeng and Mao, Dongxing and Zhang, Jiawei and Han, Weiming and Dong, Zhuobai and Li, Linjie and Lin, Yiqi and Yang, Zhengyuan and Qin, Libo and Zhang, Fuwei and others},
  journal={arXiv preprint arXiv:2502.07870},
  year={2025}
}

@article{geneval,
  title={Geneval: An object-focused framework for evaluating text-to-image alignment},
  author={Ghosh, Dhruba and Hajishirzi, Hannaneh and Schmidt, Ludwig},
  journal={Advances in Neural Information Processing Systems},
  volume={36},
  pages={52132--52152},
  year={2023}
}

@inproceedings{nhr,
  title={Nohumansrequired: Autonomous high-quality image editing triplet mining},
  author={Kuprashevich, Maksim and Alekseenko, Grigorii and Tolstykh, Irina and Fedorov, Georgii and Suleimanov, Bulat and Dokholyan, Vladimir and Gordeev, Aleksandr},
  booktitle={Proceedings of the IEEE/CVF Winter Conference on Applications of Computer Vision},
  pages={6059--6068},
  year={2026}
}

@article{unicedit,
  title={UnicEdit-10M: A Dataset and Benchmark Breaking the Scale-Quality Barrier via Unified Verification for Reasoning-Enriched Edits},
  author={Ye, Keming and Huang, Zhipeng and Fu, Canmiao and Liu, Qingyang and Cai, Jiani and Lv, Zheqi and Li, Chen and Lyu, Jing and Zhao, Zhou and Zhang, Shengyu},
  journal={arXiv preprint arXiv:2512.02790},
  year={2025}
}

@article{sharegpt4o,
  title={Sharegpt-4o-image: Aligning multimodal models with gpt-4o-level image generation},
  author={Chen, Junying and Cai, Zhenyang and Chen, Pengcheng and Chen, Shunian and Ji, Ke and Wang, Xidong and Yang, Yunjin and Wang, Benyou},
  journal={arXiv preprint arXiv:2506.18095},
  year={2025}
}

@article{pico,
  title={Pico-banana-400k: A large-scale dataset for text-guided image editing},
  author={Qian, Yusu and Bocek-Rivele, Eli and Song, Liangchen and Tong, Jialing and Yang, Yinfei and Lu, Jiasen and Hu, Wenze and Gan, Zhe},
  journal={arXiv preprint arXiv:2510.19808},
  year={2025}
}

@article{januspro,
  title={Janus-pro: Unified multimodal understanding and generation with data and model scaling},
  author={Chen, Xiaokang and Wu, Zhiyu and Liu, Xingchao and Pan, Zizheng and Liu, Wen and Xie, Zhenda and Yu, Xingkai and Ruan, Chong},
  journal={arXiv preprint arXiv:2501.17811},
  year={2025}
}

@article{anytext,
  title={Anytext: Multilingual visual text generation and editing},
  author={Tuo, Yuxiang and Xiang, Wangmeng and He, Jun-Yan and Geng, Yifeng and Xie, Xuansong},
  journal={arXiv preprint arXiv:2311.03054},
  year={2023}
}

@inproceedings{textdiffuser2,
  title={Textdiffuser-2: Unleashing the power of language models for text rendering},
  author={Chen, Jingye and Huang, Yupan and Lv, Tengchao and Cui, Lei and Chen, Qifeng and Wei, Furu},
  booktitle={European Conference on Computer Vision},
  pages={386--402},
  year={2024},
  organization={Springer}
}

@inproceedings{pixartsigma,
  title={Pixart-$\sigma$: Weak-to-strong training of diffusion transformer for 4k text-to-image generation},
  author={Chen, Junsong and Ge, Chongjian and Xie, Enze and Wu, Yue and Yao, Lewei and Ren, Xiaozhe and Wang, Zhongdao and Luo, Ping and Lu, Huchuan and Li, Zhenguo},
  booktitle={European Conference on Computer Vision},
  pages={74--91},
  year={2024},
  organization={Springer}
}

@inproceedings{infinity,
  title={Infinity: Scaling bitwise autoregressive modeling for high-resolution image synthesis},
  author={Han, Jian and Liu, Jinlai and Jiang, Yi and Yan, Bin and Zhang, Yuqi and Yuan, Zehuan and Peng, Bingyue and Liu, Xiaobing},
  booktitle={Proceedings of the Computer Vision and Pattern Recognition Conference},
  pages={15733--15744},
  year={2025}
}

@inproceedings{illume,
  title={Illume: Illuminating your llms to see, draw, and self-enhance},
  author={Wang, Chunwei and Lu, Guansong and Yang, Junwei and Huang, Runhui and Han, Jianhua and Hou, Lu and Zhang, Wei and Xu, Hang},
  booktitle={Proceedings of the IEEE/CVF International Conference on Computer Vision},
  pages={21612--21622},
  year={2025}
}

@inproceedings{ddit,
  title={Dual diffusion for unified image generation and understanding},
  author={Li, Zijie and Li, Henry and Shi, Yichun and Farimani, Amir Barati and Kluger, Yuval and Yang, Linjie and Wang, Peng},
  booktitle={Proceedings of the Computer Vision and Pattern Recognition Conference},
  pages={2779--2790},
  year={2025}
}

@article{metaqueryxl,
  title={Transfer between modalities with metaqueries},
  author={Pan, Xichen and Shukla, Satya Narayan and Singh, Aashu and Zhao, Zhuokai and Mishra, Shlok Kumar and Wang, Jialiang and Xu, Zhiyang and Chen, Jiuhai and Li, Kunpeng and Juefei-Xu, Felix and others},
  journal={arXiv preprint arXiv:2504.06256},
  year={2025}
}

@article{dalle2,
  title={Hierarchical text-conditional image generation with clip latents},
  author={Ramesh, Aditya and Dhariwal, Prafulla and Nichol, Alex and Chu, Casey and Chen, Mark},
  journal={arXiv preprint arXiv:2204.06125},
  volume={1},
  number={2},
  pages={3},
  year={2022}
}

@article{cogview,
  title={Cogview: Mastering text-to-image generation via transformers},
  author={Ding, Ming and Yang, Zhuoyi and Hong, Wenyi and Zheng, Wendi and Zhou, Chang and Yin, Da and Lin, Junyang and Zou, Xu and Shao, Zhou and Yang, Hongxia and others},
  journal={Advances in neural information processing systems},
  volume={34},
  pages={19822--19835},
  year={2021}
}

@article{vqvae,
  title={Neural discrete representation learning},
  author={Van Den Oord, Aaron and Vinyals, Oriol and others},
  journal={Advances in neural information processing systems},
  volume={30},
  year={2017}
}

@inproceedings{vqgan,
  title={Taming transformers for high-resolution image synthesis},
  author={Esser, Patrick and Rombach, Robin and Ommer, Bjorn},
  booktitle={Proceedings of the IEEE/CVF conference on computer vision and pattern recognition},
  pages={12873--12883},
  year={2021}
}

@article{parti,
  title={Scaling autoregressive models for content-rich text-to-image generation},
  author={Yu, Jiahui and Xu, Yuanzhong and Koh, Jing Yu and Luong, Thang and Baid, Gunjan and Wang, Zirui and Vasudevan, Vijay and Ku, Alexander and Yang, Yinfei and Ayan, Burcu Karagol and others},
  journal={arXiv preprint arXiv:2206.10789},
  volume={2},
  number={3},
  pages={5},
  year={2022}
}

@article{var,
  title={Visual autoregressive modeling: Scalable image generation via next-scale prediction},
  author={Tian, Keyu and Jiang, Yi and Yuan, Zehuan and Peng, Bingyue and Wang, Liwei},
  journal={Advances in neural information processing systems},
  volume={37},
  pages={84839--84865},
  year={2024}
}

@article{mar,
  title={Autoregressive image generation without vector quantization},
  author={Li, Tianhong and Tian, Yonglong and Li, He and Deng, Mingyang and He, Kaiming},
  journal={Advances in Neural Information Processing Systems},
  volume={37},
  pages={56424--56445},
  year={2024}
}

@article{ddpm,
  title={Denoising diffusion probabilistic models},
  author={Ho, Jonathan and Jain, Ajay and Abbeel, Pieter},
  journal={Advances in neural information processing systems},
  volume={33},
  pages={6840--6851},
  year={2020}
}

@article{imagen,
  title={Photorealistic text-to-image diffusion models with deep language understanding},
  author={Saharia, Chitwan and Chan, William and Saxena, Saurabh and Li, Lala and Whang, Jay and Denton, Emily L and Ghasemipour, Kamyar and Gontijo Lopes, Raphael and Karagol Ayan, Burcu and Salimans, Tim and others},
  journal={Advances in neural information processing systems},
  volume={35},
  pages={36479--36494},
  year={2022}
}

@article{dinov2,
  title={Dinov2: Learning robust visual features without supervision},
  author={Oquab, Maxime and Darcet, Timoth{\'e}e and Moutakanni, Th{\'e}o and Vo, Huy and Szafraniec, Marc and Khalidov, Vasil and Fernandez, Pierre and Haziza, Daniel and Massa, Francisco and El-Nouby, Alaaeldin and others},
  journal={arXiv preprint arXiv:2304.07193},
  year={2023}
}

@article{prompt2prompt,
  title={Prompt-to-prompt image editing with cross attention control},
  author={Hertz, Amir and Mokady, Ron and Tenenbaum, Jay and Aberman, Kfir and Pritch, Yael and Cohen-Or, Daniel},
  journal={arXiv preprint arXiv:2208.01626},
  year={2022}
}

@inproceedings{nulltext,
  title={Null-text inversion for editing real images using guided diffusion models},
  author={Mokady, Ron and Hertz, Amir and Aberman, Kfir and Pritch, Yael and Cohen-Or, Daniel},
  booktitle={Proceedings of the IEEE/CVF conference on computer vision and pattern recognition},
  pages={6038--6047},
  year={2023}
}

@inproceedings{smartedit,
  title={Smartedit: Exploring complex instruction-based image editing with multimodal large language models},
  author={Huang, Yuzhou and Xie, Liangbin and Wang, Xintao and Yuan, Ziyang and Cun, Xiaodong and Ge, Yixiao and Zhou, Jiantao and Dong, Chao and Huang, Rui and Zhang, Ruimao and others},
  booktitle={Proceedings of the IEEE/CVF Conference on Computer Vision and Pattern Recognition},
  pages={8362--8371},
  year={2024}
}

@article{genaibench,
  title={Genai-bench: Evaluating and improving compositional text-to-visual generation},
  author={Li, Baiqi and Lin, Zhiqiu and Pathak, Deepak and Li, Jiayao and Fei, Yixin and Wu, Kewen and Ling, Tiffany and Xia, Xide and Zhang, Pengchuan and Neubig, Graham and others},
  journal={arXiv preprint arXiv:2406.13743},
  year={2024}
}

@inproceedings{dsg,
  title={Davidsonian scene graph: Improving reliability in fine-grained evaluation for text-to-image generation},
  author={Cho, Jaemin and Hu, Yushi and Baldridge, Jason and Garg, Roopal and Anderson, Peter and Krishna, Ranjay and Bansal, Mohit and Pont-Tuset, Jordi and Wang, Su},
  booktitle={International conference on learning representations},
  volume={2024},
  pages={15625--15645},
  year={2024}
}

@article{i2ebench,
  title={I2ebench: A comprehensive benchmark for instruction-based image editing},
  author={Ma, Yiwei and Ji, Jiayi and Ye, Ke and Lin, Weihuang and Wang, Zhibin and Zheng, Yonghan and Zhou, Qiang and Sun, Xiaoshuai and Ji, Rongrong},
  journal={Advances in Neural Information Processing Systems},
  volume={37},
  pages={41494--41516},
  year={2024}
}

@inproceedings{sdxl,
  title={Sdxl: Improving latent diffusion models for high-resolution image synthesis},
  author={Podell, Dustin and English, Zion and Lacey, Kyle and Blattmann, Andreas and Dockhorn, Tim and M{\"u}ller, Jonas and Penna, Joe and Rombach, Robin},
  booktitle={International Conference on Learning Representations},
  volume={2024},
  pages={1862--1874},
  year={2024}
}

@article{t2ir1,
  title={T2i-r1: Reinforcing image generation with collaborative semantic-level and token-level cot},
  author={Jiang, Dongzhi and Guo, Ziyu and Zhang, Renrui and Zong, Zhuofan and Li, Hao and Zhuo, Le and Yan, Shilin and Heng, Pheng-Ann and Li, Hongsheng},
  journal={Advances in Neural Information Processing Systems},
  volume={38},
  pages={39856--39890},
  year={2026}
}

@article{laion5b,
  title={Laion-5b: An open large-scale dataset for training next generation image-text models},
  author={Schuhmann, Christoph and Beaumont, Romain and Vencu, Richard and Gordon, Cade and Wightman, Ross and Cherti, Mehdi and Coombes, Theo and Katta, Aarush and Mullis, Clayton and Wortsman, Mitchell and others},
  journal={Advances in neural information processing systems},
  volume={35},
  pages={25278--25294},
  year={2022}
}

\newpage
\appendix

\section{Full Class-to-Image Results}
\label{app:c2i}

Tables~\ref{tab:full_llamagen} and~\ref{tab:full_transfusion} report the complete evaluation metrics at more checkpoints for the class-to-image experiments.

\begin{table}[h]
\caption{\textbf{LlamaGen results on ImageNet 512$\times$512.}}
\label{tab:full_llamagen}
\centering
\small
\setlength{\tabcolsep}{3.5pt}
\begin{tabular}{c cccc cccc}
\toprule
& \multicolumn{4}{c}{LlamaGen\textsubscript{\scriptsize +\method{}}} & \multicolumn{4}{c}{LlamaGen} \\
\cmidrule(lr){2-5} \cmidrule(lr){6-9}
Epoch & FID$\downarrow$ & IS$\uparrow$ & Prec$\uparrow$ & Rec$\uparrow$ & FID$\downarrow$ & IS$\uparrow$ & Prec$\uparrow$ & Rec$\uparrow$ \\
\midrule
30 & 38.68 & 34.1 & 0.453 & 0.611 & 28.32 & 40.8 & 0.480 & 0.647 \\
40 & 28.79 & 45.7 & 0.502 & 0.623 & 28.43 & 40.8 & 0.479 & 0.655 \\
50 & 22.07 & 53.7 & 0.520 & 0.632 & 29.31 & 39.7 & 0.472 & 0.663 \\
60 & 14.12 & 74.5 & 0.584 & 0.621 & 26.63 & 42.5 & 0.484 & 0.660 \\
70 & 10.91 & 82.4 & 0.598 & 0.626 & 24.89 & 44.2 & 0.497 & 0.660 \\
80 & 9.35 & 88.8 & 0.614 & 0.629 & 25.16 & 44.4 & 0.492 & 0.667 \\
90 & 9.50 & 92.1 & 0.611 & 0.625 & 25.74 & 43.9 & 0.497 & 0.658 \\
100 & 8.69 & 95.7 & 0.616 & 0.624 & 24.56 & 45.1 & 0.498 & 0.666 \\
\bottomrule
\end{tabular}
\end{table}

\begin{table}[h]
\caption{\textbf{Transfusion results on ImageNet 256$\times$256.} Steps 72k--130k are shared training before the fork point. ``---'' indicates the model variant does not exist at that step.}
\label{tab:full_transfusion}
\centering
\small
\setlength{\tabcolsep}{3.5pt}
\begin{tabular}{c cccc cccc}
\toprule
& \multicolumn{4}{c}{Transfusion\textsubscript{\scriptsize +\method{}}} & \multicolumn{4}{c}{Transfusion} \\
\cmidrule(lr){2-5} \cmidrule(lr){6-9}
Step (k) & FID$\downarrow$ & IS$\uparrow$ & Prec$\uparrow$ & Rec$\uparrow$ & FID$\downarrow$ & IS$\uparrow$ & Prec$\uparrow$ & Rec$\uparrow$ \\
\midrule
72 & --- & --- & --- & --- & 29.05 & 45.9 & 0.421 & 0.614 \\
96 & --- & --- & --- & --- & 22.07 & 59.3 & 0.468 & 0.621 \\
130 & --- & --- & --- & --- & 18.26 & 69.6 & 0.488 & 0.629 \\
\midrule
140 & 198.92 & 4.1 & 0.084 & 0.376 & 17.91 & 69.8 & 0.492 & 0.629 \\
150 & 36.62 & 57.5 & 0.518 & 0.354 & 17.73 & 70.1 & 0.495 & 0.626 \\
162 & 17.10 & 147.8 & 0.676 & 0.271 & 17.66 & 70.2 & 0.496 & 0.628 \\
172 & 15.32 & 166.5 & 0.689 & 0.280 & 17.00 & 72.4 & 0.501 & 0.629 \\
184 & 14.02 & 172.0 & 0.691 & 0.307 & 16.72 & 73.4 & 0.497 & 0.632 \\
206 & 12.72 & 166.1 & 0.679 & 0.343 & 15.50 & 77.5 & 0.511 & 0.633 \\
226 & 12.12 & 159.6 & 0.672 & 0.369 & 15.00 & 80.2 & 0.513 & 0.628 \\
246 & 11.87 & 152.9 & 0.664 & 0.388 & 15.26 & 79.0 & 0.509 & 0.629 \\
266 & 11.43 & 148.2 & 0.660 & 0.404 & 15.18 & 78.8 & 0.508 & 0.637 \\
330 & 10.92 & 136.8 & 0.641 & 0.435 & --- & --- & --- & --- \\
530 & 12.10 & 107.7 & 0.593 & 0.495 & --- & --- & --- & --- \\
\bottomrule
\end{tabular}
\end{table}

\section{Text Rendering at Matched Steps}
\label{app:textatlas_72k}

To address potential concerns about different numbers of gradient updates at evaluation, we also compare Show-o2\textsubscript{\scriptsize +\method{}} and Show-o2$^*$ at the same checkpoint step (72k). Because the SigLIP 2 tokens increase the per-sample sequence length, Show-o2\textsubscript{\scriptsize +\method{}} uses a smaller micro batch size (8 vs.\ 12) to maintain comparable per-step token throughput ($8 \times 1{,}856 = 14{,}848$ vs.\ $12 \times 1{,}280 = 15{,}360$ tokens/step, $<3.5\%$ difference). This means that at the same step count, Show-o2\textsubscript{\scriptsize +\method{}} has seen fewer total images and fewer image tokens. Despite this data disadvantage, it still achieves substantially higher OCR accuracy (Table~\ref{tab:textatlas_72k}), confirming that the improvement stems from \method{}.

\begin{table}[h]
\caption{\textbf{Text rendering at matched steps (72k).} All results at CFG$=$5.5. $*$: continued training on 6.35M text rendering data.}
\label{tab:textatlas_72k}
\centering
\small
\setlength{\tabcolsep}{3pt}
\begin{tabular}{lccccccccccccc}
\toprule
& \multicolumn{4}{c}{TextScenesHQ} && \multicolumn{4}{c}{TextVisionBlend} && \multicolumn{3}{c}{StyleTextSynth} \\
\cmidrule(lr){2-5} \cmidrule(lr){7-10} \cmidrule(lr){12-14}
Method & CS$\uparrow$ & Acc$\uparrow$ & F1$\uparrow$ & CER$\downarrow$ && CS$\uparrow$ & Acc$\uparrow$ & F1$\uparrow$ & CER$\downarrow$ && Acc$\uparrow$ & F1$\uparrow$ & CER$\downarrow$ \\
\midrule
Show-o2$^*$ (72k) & \textbf{0.29} & 2.62 & 4.36 & 0.88 && 0.16 & 76.45 & 77.40 & 0.22 && 21.83 & 27.72 & \textbf{0.59} \\
\rowcolor{tabhighlight} \textbf{Show-o2\textsubscript{\scriptsize +\method{}} (ours, 72k)} & 0.26 & \textbf{12.10} & \textbf{20.66} & \textbf{0.82} && 0.16 & \textbf{79.83} & \textbf{82.34} & \textbf{0.20} && \textbf{26.73} & \textbf{37.05} & 0.67 \\
\bottomrule
\end{tabular}
\end{table}

\section{Internal vs.\ External Training Details}
\label{app:training}

\paragraph{SigLIP 2 encoder.}
We use the SigLIP 2-Giant-Patch16-384 model~\cite{siglip2} as the visual semantic encoder. Target images are resized to $384 \times 384$ with white padding to extract $24 \times 24 = 576$ SigLIP 2 tokens. The resulting tokens are then discretized using the frozen SigLIP-VQ tokenizer from X-Omni~\cite{xomni}, which adopts a codebook of 16,384 entries.

\begin{table}[h]
\caption{\textbf{Training hyperparameters for each stage.} Stage 1 is shared, Stages 2 and 3 differ between internal and external architectures. ``CE only'' indicates that only cross-entropy loss on SigLIP 2 tokens is used (no flow matching loss). ``Flow only'' indicates that only flow matching loss is used (no cross-entropy loss).}
\label{tab:training_details}
\centering
\small
\setlength{\tabcolsep}{3.5pt}
\begin{tabular}{lccccc}
\toprule
& & \multicolumn{2}{c}{Stage 2 (T2I)} & \multicolumn{2}{c}{Stage 3 (Editing)} \\
\cmidrule(lr){3-4} \cmidrule(lr){5-6}
& Stage 1 & Internal & External & Internal & External \\
\midrule
GPUs & 16 H200 & 24 H200 & 24 H200 & 24 H200 & 24 H200 \\
Data & 39M & 39M & 39M & 5.4M & 5.4M \\
Epochs & 4 & 2 & 2 & $\sim$1.5 & $\sim$1.5 \\
BS / GPU & 30 & 8 & 8 & 5 & 5 \\
Grad Accum & 1 & 1 & 1 & 2 & 2 \\
LR & $10^{-4}$ & $10^{-4}$ & $10^{-4}$ & $5 \times 10^{-5}$ & $5 \times 10^{-5}$ \\
LR Schedule & const+warm & const+warm & const+warm & const+warm & const+warm \\
EMA & --- & 0.9995 & 0.9995 & 0.9995 & 0.9995 \\
$\lambda_{\text{CE}}:\lambda_{\text{flow}}$ & CE only & 0.03$\to$0.1:1.0 & Flow only & 0.1:1.0 & CE only \\
Frozen & DiT & --- & AR & --- & DiT \\
\bottomrule
\end{tabular}
\end{table}

\paragraph{Stage 1: SigLIP 2 alignment.}
Both architectures share the same Stage-1 checkpoint. The model learns to autoregressively predict SigLIP 2 tokens given text prompts, using only cross-entropy loss. The DiT and all image-related modules are frozen, only the AR (LLM) is trained. We train for 4 epochs on 39M mixed data (32M T2I + 6.35M text rendering) with constant LR $10^{-4}$ and 2,000 warmup steps.

\paragraph{Stage 2: T2I training.}
Both Internal\textsubscript{\scriptsize +\method{}} and External\textsubscript{\scriptsize +\method{}} load the Stage-1 checkpoint and are trained on 39M mixed data for 2 epochs. The two training strategies follow the standard pipelines of their respective architectures. Internal\textsubscript{\scriptsize +\method{}}: all parameters are unfrozen and jointly trained following the Transfusion-style training paradigm. For Internal\textsubscript{\scriptsize +\method{}}, the CE loss coefficient $\lambda_{\text{CE}}$ is progressively increased from 0.03 to 0.1 during training to balance the CE and flow matching objectives. External\textsubscript{\scriptsize +\method{}}: the AR is frozen throughout and only the DiT is trained for the full 2 epochs, following the standard external paradigm~\cite{xomni,blip3o} where the AR serves as a fixed semantic encoder, so no loss balancing is needed.

\paragraph{Stage 3: Editing.}
Both models load their respective Stage-2 T2I checkpoints and add a reference image encoder initialized from Show-o2-1.5B~\cite{showo2} weights, increasing parameters from 4.16B to 4.57B. Both are trained on 5.4M editing pairs for approximately 1.5 epochs (70k steps) with the same data, batch size, and training duration.

\textit{Internal\textsubscript{\scriptsize +\method{}}}: all parameters are unfrozen. The model receives the reference image, edit instruction, and SigLIP 2 tokens within a single sequence, and is trained with both cross-entropy loss ($\lambda_{\text{CE}} = 0.1$) and flow matching loss ($\lambda_{\text{flow}} = 1.0$). We adopt InstructPix2Pix-style~\cite{instructpix2pix} conditional dropout: with 5\% probability drop the reference image, with 5\% probability drop text and SigLIP 2 tokens jointly, and with 5\% probability drop all conditions. Text and SigLIP 2 tokens are coupled because both serve as semantic prompts for the generation process (details in Appendix~\ref{app:editing_cfg}).

\textit{External\textsubscript{\scriptsize +\method{}}}: the DiT is frozen and only the AR model is trained. In the external architecture, the DiT always receives SigLIP 2 features as conditioning regardless of the task. For editing, the new capability the model needs to learn is how to incorporate the reference image information into the SigLIP 2 representation, which is handled entirely by the AR component. Therefore, only the AR is trained with cross-entropy loss. Conditional dropout uses the same three mutually exclusive 5\% cases as the internal model (no reference, no text and SigLIP 2, or null) (details in Appendix~\ref{app:editing_cfg}).

\section{Editing CFG Formulation}
\label{app:editing_cfg}

\paragraph{Internal model (Internal\textsubscript{\scriptsize +\method{}}).}
We denote the reference image tokens as $\mathbf{x}_{\text{ref}}$, the edit instruction as $t$, the SigLIP 2 visual prompt tokens as $\mathbf{s}$, and the generated image tokens as $\mathbf{x}_{\text{gen}}$. Special tokens include $[\textsc{bos}]$/$[\textsc{eos}]$ (sequence boundaries), $[\textsc{boi}]$/$[\textsc{eoi}]$ (image boundaries), and $[\textsc{bot}]$/$[\textsc{eot}]$ (visual prompt boundaries). Same notations apply to the external model. The internal model uses four input configurations during training, with the last three at 5\% probability each:
\begin{align}
&\text{Full:} \quad [\textsc{bos}][\textsc{boi}]\;\mathbf{x}_{\text{ref}}\;[\textsc{eoi}]\;t\;[\textsc{bot}]\;\mathbf{s}\;[\textsc{eot}]\;[\textsc{boi}]\;\mathbf{x}_{\text{gen}}\;[\textsc{eoi}][\textsc{eos}] \nonumber \\
&\text{No-text:} \quad [\textsc{bos}][\textsc{boi}]\;\mathbf{x}_{\text{ref}}\;[\textsc{eoi}]\;[\textsc{boi}]\;\mathbf{x}_{\text{gen}}\;[\textsc{eoi}][\textsc{eos}] \nonumber \\
&\text{No-ref:} \quad [\textsc{bos}]\;t\;[\textsc{bot}]\;\mathbf{s}\;[\textsc{eot}]\;[\textsc{boi}]\;\mathbf{x}_{\text{gen}}\;[\textsc{eoi}][\textsc{eos}] \nonumber \\
&\text{Null:} \quad [\textsc{bos}][\textsc{boi}]\;\mathbf{x}_{\text{gen}}\;[\textsc{eoi}][\textsc{eos}] \nonumber
\end{align}
Note that text and SigLIP 2 tokens are always dropped \textit{jointly} in the internal model, since they are semantically coupled through the AR model.

\paragraph{External model (External\textsubscript{\scriptsize +\method{}}).}
The external model's AR component uses four input configurations during editing training, with the last three at 5\% probability each:

\begin{align}
&\text{Full:} \quad [\textsc{bos}][\textsc{boi}]\;\mathbf{x}_{\text{ref}}\;[\textsc{eoi}]\;t\;[\textsc{bot}]\;\mathbf{s}\;[\textsc{eot}][\textsc{eos}] \nonumber \\
&\text{No-text:} \quad [\textsc{bos}][\textsc{boi}]\;\mathbf{x}_{\text{ref}}\;[\textsc{eoi}][\textsc{eos}] \nonumber \\
&\text{No-ref:} \quad [\textsc{bos}]\;t\;[\textsc{bot}]\;\mathbf{s}\;[\textsc{eot}][\textsc{eos}] \nonumber \\
&\text{Null:} \quad [\textsc{bos}][\textsc{eos}] \nonumber
\end{align}

 As in the internal model, the four cases are mutually exclusive (a single random draw selects one of them), and text and SigLIP 2 tokens are always dropped jointly. The DiT is conditioned on the AR hidden states at the SigLIP 2 positions. When these tokens are absent (No-text and Null), it receives an all-zero SigLIP 2 feature. During editing, only the AR component is trained (DiT remains frozen), since the AR effectively serves as a visual semantic encoder that translates the reference image and edit instruction into SigLIP 2 tokens for the DiT, following the standard external paradigm~\cite{xomni,blip3o}.

\paragraph{Inference CFG.} Both models generate the visual prompt $\mathbf{s}$ greedily, without guidance, and then combine three velocity predictions at each denoising step following InstructPix2Pix~\cite{instructpix2pix}:
\begin{equation*}
\hat{\mathbf{v}} = \mathbf{v}_{\varnothing} + s_I\,(\mathbf{v}_{\text{ref}} - \mathbf{v}_{\varnothing}) + s_T\,(\mathbf{v}_{\text{full}} - \mathbf{v}_{\text{ref}}),
\end{equation*}
where $\mathbf{v}_{\text{full}}$, $\mathbf{v}_{\text{ref}}$, and $\mathbf{v}_{\varnothing}$ are predicted under the Full, No-text, and Null configurations, with $s_I{=}1.0$ and $s_T{=}2.5$. For the external model, the No-text and Null configurations both give the DiT an all-zero SigLIP 2 feature, so $\mathbf{v}_{\text{ref}}{=}\mathbf{v}_{\varnothing}$ and the rule reduces to standard two-way guidance $\mathbf{v}_{\varnothing} + s_T(\mathbf{v}_{\text{full}} - \mathbf{v}_{\varnothing})$.

\section{Training Loss Curves}
\label{app:loss}

\begin{figure}[h]
\centering
\includegraphics[width=\linewidth]{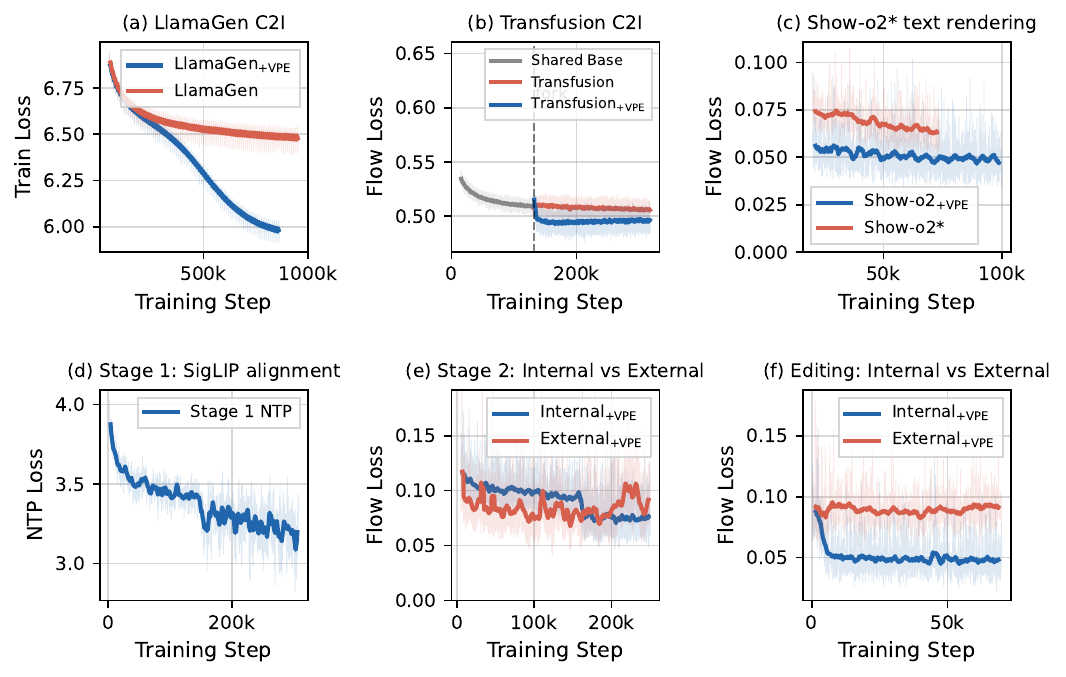}
\caption{\textbf{Training loss curves.} \textit{Top row:} \method{} vs.\ baseline training loss for each framework. \textit{(a)} LlamaGen\textsubscript{\scriptsize +\method{}} converges to lower loss than LlamaGen. \textit{(b)} Transfusion\textsubscript{\scriptsize +\method{}} converges to lower flow loss than Transfusion after forking from the shared base. \textit{(c)} Show-o2\textsubscript{\scriptsize +\method{}} achieves lower flow loss than Show-o2$^*$ on text rendering. \textit{Bottom row:} Internal vs.\ External architecture training. \textit{(d)} Stage-1 SigLIP 2 alignment. \textit{(e)} Stage-2: both architectures achieve similar flow loss. \textit{(f)} Editing: Internal\textsubscript{\scriptsize +\method{}} achieves lower flow loss than External\textsubscript{\scriptsize +\method{}}, consistent with its superior preservation metrics (Table~\ref{tab:editing}).}
\label{fig:loss_all}
\end{figure}

We provide training loss curves for all experiments in Figure~\ref{fig:loss_all}. Smooth curves (moving average) are overlaid on raw data (low opacity) for clarity.

\section{Additional Results}
\label{app:extra}

\subsection{Training and inference efficiency}
\label{app:efficiency}

\begin{figure}[h]
\centering
\includegraphics[width=\linewidth]{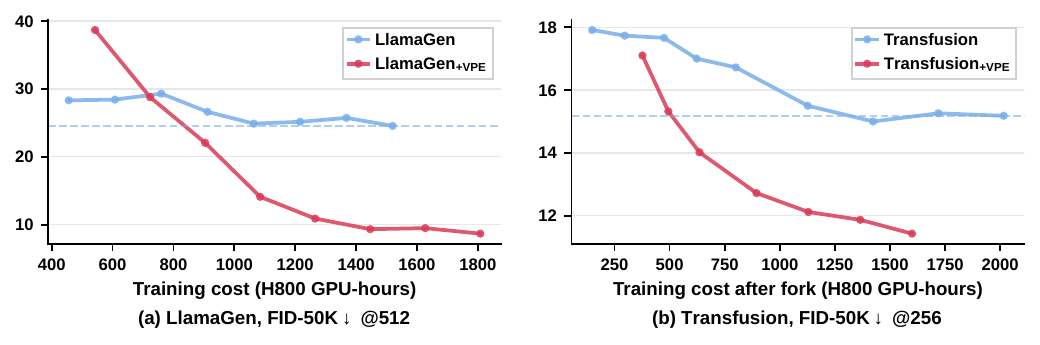}
\caption{\textbf{Quality vs.\ training compute.} FID-50K against cumulative training GPU-hours. Dashed lines mark the baseline's final FID. For Transfusion, cost is counted from the 130k-step fork.}
\label{fig:qvc}
\end{figure}

Table~\ref{tab:train_cost} reports training cost and Table~\ref{tab:infer_cost} inference cost. C2I experiments run on H800 GPUs and text rendering on H200 GPUs. LlamaGen\textsubscript{\scriptsize +\method{}} incurs an 18.9\% per-step overhead from its longer sequences, but reaches the baseline's final FID (24.56) by epoch 50 ($\sim$904 H800 GPU-hours vs.\ 1,520), a 40\% reduction in total compute to match quality. Under fixed 10,240-token packing, Transfusion\textsubscript{\scriptsize +\method{}} is 20.6\% cheaper per run, and Show-o2\textsubscript{\scriptsize +\method{}} is compute-neutral ($<2\%$ difference). Figure~\ref{fig:qvc} plots FID against training GPU-hours. At inference, AR planning adds a one-time per-image cost of 2--8\,s, depending on the number of visual prompt tokens (Table~\ref{tab:infer_cost}).

\begin{table}[h]
\caption{\textbf{Training cost.} Overhead is relative to the baseline of the same framework. For Transfusion, cost covers the 136k steps after the fork.}
\label{tab:train_cost}
\centering
\small
\setlength{\tabcolsep}{3pt}
\begin{tabular}{lcccccc}
\toprule
Setting & Seq.\ len & Global batch & Steps & GPU-hours & Overhead & Peak mem \\
\midrule
LlamaGen & 1,025 & 128 & 1,000,000 & 1,520 (H800) & --- & 37.7\,GiB \\
LlamaGen\textsubscript{\scriptsize +\method{}} & 1,221 & 128 & 1,000,000 & 1,808 (H800) & $+$18.9\% & 43.8\,GiB \\
\midrule
Transfusion & 10,240 (packed) & 32 & 136,000 & 2,016 (H800) & --- & 6.1\,GiB \\
Transfusion\textsubscript{\scriptsize +\method{}} & 10,240 (packed) & 32 & 136,000 & 1,600 (H800) & $-$20.6\% & 4.51\,GiB \\
\midrule
Show-o2$^*$ & 1,280 & 288 & 70,000 & 2,224 (H200) & --- & 120.8\,GiB \\
Show-o2\textsubscript{\scriptsize +\method{}} & 1,856 & 192 & 70,000 & 2,186 (H200) & $\sim-$1.7\% & 120.4\,GiB \\
\bottomrule
\end{tabular}
\end{table}

\begin{table}[h]
\caption{\textbf{Inference cost} (bs$=$1, 50 timesteps).}
\label{tab:infer_cost}
\centering
\small
\begin{tabular}{lccc}
\toprule
Setting & AR planning & Generation & Peak memory \\
\midrule
LlamaGen & --- & 16.02\,s & 2.71\,GiB \\
LlamaGen\textsubscript{\scriptsize +\method{}} & 196 tokens / 3.04\,s & 16.34\,s & 2.99\,GiB \\
\midrule
Transfusion & --- & 1.21\,s & 2.50\,GiB \\
Transfusion\textsubscript{\scriptsize +\method{}} & 121 tokens / 2.40\,s & 1.17\,s & 2.51\,GiB \\
\midrule
Show-o2$^*$ & --- & 3.93\,s & 7.69\,GiB \\
Show-o2\textsubscript{\scriptsize +\method{}} & 576 tokens / 8.28\,s & 4.06\,s & 7.80\,GiB \\
\bottomrule
\end{tabular}
\end{table}

\subsection{Internal\textsubscript{\scriptsize +\method{}} inference procedure}
\label{app:inference}
Algorithm~\ref{alg:internal} summarizes inference for Internal\textsubscript{\scriptsize +\method{}} (Section~\ref{sec:arch}), taking image editing as the example. Inference runs in two steps: the AR expert first decodes the visual prompt with a KV cache, and the DiT tokens then attend to the AR keys/values and to themselves at every denoising step. For text-to-image generation, the procedure is identical except that there is no reference image $\mathbf{x}_{\text{ref}}$ in the input. Classifier-free guidance is omitted from the algorithm for clarity.
\begin{algorithm}[h]
\caption{Inference procedure of Internal\textsubscript{\scriptsize +\method{}} for image editing.}
\label{alg:internal}
\small
\begin{algorithmic}[1]
\State $\mathbf{s} \gets \textsc{AR.Generate}([\mathbf{x}_{\text{ref}}, t])$ \Comment{Step A: causal decoding with KV cache}
\State $\mathbf{KV}_{\text{AR}} \gets$ keys/values of $[\mathbf{x}_{\text{ref}}, t, \mathbf{s}]$ at every layer \Comment{fixed during Step B}
\State $\mathbf{z} \sim \mathcal{N}(\mathbf{0}, \mathbf{I})$
\For{$k = 1, \dots, K$} \Comment{Step B: denoising}
  \State $\mathbf{v} \gets \textsc{DiT}(\mathbf{z}, \tau_k;\, \mathbf{KV}_{\text{AR}})$ \Comment{DiT queries attend to $[\mathbf{KV}_{\text{AR}}; \mathbf{KV}_{\text{DiT}}]$ (Eq.~\ref{eq:mot_attn})}
  \State $\mathbf{z} \gets \mathbf{z} + (\tau_{k+1} - \tau_k)\,\mathbf{v}$
\EndFor
\State \Return $\textsc{VAE.Decode}(\mathbf{z})$
\end{algorithmic}
\end{algorithm}

\subsection{Progressive schedule ablation}
\label{app:schedule}

Table~\ref{tab:schedule_ablation} compares the progressive schedules of Section~\ref{sec:schedule} with fixed alternatives. The progressive schedule gives the best final FID in both frameworks. For LlamaGen, fully exposed prompts perform worst early (23.10 at epoch 50): the model relies heavily on SigLIP 2 tokens while its planning ability is still poor. The gap narrows as planning improves (10.42 at epoch 80), consistent with the shrinking ground-truth gap in Table~\ref{tab:gt_vp}. Fixed 0.5 dropout is more robust early (20.41 at epoch 50) because the model keeps a direct conditioning pathway, but it under-uses the semantic plan later and caps the final quality. The progressive schedule balances the two: early masking builds robustness to imperfect plans, and late unmasking lets the model fully use the plan. For Transfusion, a fixed $\lambda_{\text{CE}}{=}0.1$ gives the worst early FID among the loss-weighting variants (17.81 at 162k), as the CE gradient dominates flow-matching learning. A fixed $\lambda_{\text{CE}}{=}0.03$ dilutes the CE signal in mid-training and limits plan quality. Token dropout with $\lambda_{\text{CE}}{=}0.1$ gives the worst final FID (15.68), as the masking strategy that works for discrete tokens is unsuitable for continuous sequences.

\begin{table}[h]
\caption{\textbf{Progressive schedule ablation} (FID-50K$\downarrow$, no CFG). All variants share the same visual prompts, data, and compute, and differ only in the schedule.}
\label{tab:schedule_ablation}
\centering
\small
\setlength{\tabcolsep}{3pt}
\begin{minipage}[t]{0.43\textwidth}
\centering
\textit{(a) LlamaGen ($p_{\text{mask}}$)}\\[2pt]
\begin{tabular}{ccccc}
\toprule
Ep & Base & \makecell{Prog.\\0.95$\to$0.05} & \makecell{Fully\\exposed} & \makecell{Fixed\\0.5} \\
\midrule
50 & 29.31 & 22.07 & 23.10 & \textbf{20.41} \\
60 & 26.63 & \textbf{14.12} & 16.01 & 16.92 \\
70 & 24.89 & \textbf{10.91} & 11.14 & 13.81 \\
80 & 25.16 & \textbf{9.35} & 10.42 & 10.26 \\
\bottomrule
\end{tabular}
\end{minipage}%
\hfill
\begin{minipage}[t]{0.55\textwidth}
\centering
\textit{(b) Transfusion ($\lambda_{\text{CE}}$)}\\[2pt]
\begin{tabular}{cccccc}
\toprule
Step & Base & \makecell{Prog.\\0.03$\to$0.1} & \makecell{Fixed\\0.1} & \makecell{Fixed\\0.03} & \makecell{0.1 +\\tok. drop} \\
\midrule
162k & 17.66 & \textbf{17.10} & 17.81 & 17.21 & 23.42 \\
184k & 16.72 & \textbf{14.02} & 14.41 & 15.45 & 19.37 \\
226k & 15.00 & \textbf{12.12} & 14.66 & 13.13 & 16.70 \\
266k & 15.18 & \textbf{11.43} & 14.78 & 12.49 & 15.68 \\
\bottomrule
\end{tabular}
\end{minipage}
\end{table}

\subsection{Ground-truth vs.\ predicted visual prompts}
\label{app:gt_vp}

We teacher-force the ground-truth SigLIP-VQ tokens of ImageNet validation images at inference to measure the gap between predicted and oracle plans (Table~\ref{tab:gt_vp}). Ground-truth prompts consistently lower FID in both frameworks, and the gap shrinks as training proceeds (LlamaGen: $-7.62$ at epoch 50 to $-5.02$ at epoch 80, and Transfusion: $-7.46$ at 162k to $-3.56$ at 266k). The gain of \method{} therefore comes from the semantic plan and grows with the quality of the predicted plan.

\begin{table}[h]
\caption{\textbf{Ground-truth vs.\ predicted visual prompts} (FID-50K$\downarrow$, no CFG).}
\label{tab:gt_vp}
\centering
\small
\begin{tabular}{lcccc}
\toprule
& Baseline & \method{} (predicted) & \method{} (GT) & $\Delta$ (GT $-$ pred) \\
\midrule
\multicolumn{5}{l}{\textit{LlamaGen}} \\
Epoch 50 & 29.31 & 22.07 & 14.45 & $-$7.62 \\
Epoch 60 & 26.63 & 14.12 & 7.28 & $-$6.84 \\
Epoch 70 & 24.89 & 10.91 & 5.57 & $-$5.34 \\
Epoch 80 & 25.16 & 9.35 & 4.33 & $-$5.02 \\
\midrule
\multicolumn{5}{l}{\textit{Transfusion}} \\
162k & 17.66 & 17.10 & 9.64 & $-$7.46 \\
184k & 16.72 & 14.02 & 9.98 & $-$4.04 \\
226k & 15.00 & 12.12 & 8.30 & $-$3.82 \\
266k & 15.18 & 11.43 & 7.87 & $-$3.56 \\
\bottomrule
\end{tabular}
\end{table}

\subsection{Controlled discrete vs.\ continuous comparison}
\label{app:disc_cont}

Our main C2I experiments validate \method{} within matched pairs in two different frameworks, so they do not compare token types directly. We therefore run a controlled experiment in the same Transfusion framework, switching only the image token type. Continuous-specific modules (timestep embedding, latent projection) are replaced by a discrete codebook embedding, giving comparable sizes (continuous 0.37B/0.39B, discrete 0.38B/0.40B for baseline/\method{}). The discrete variant uses the LlamaGen VQ-VAE~\cite{llamagen} ($16\times$ downsampling, codebook 16,384, rFID 2.19, PSNR 20.79\,dB) and the continuous variant the FLUX VAE (rFID 0.13, PSNR 30.20\,dB), both with 256 tokens at $256{\times}256$. Both are initialized from Qwen2.5-0.5B-Instruct (excluding embeddings), use packed sequences of 10,240 tokens, and train 130k shared steps plus 136k branched steps, with progressive masking (discrete) or progressive loss weights (continuous). \method{} improves both token types (Table~\ref{tab:disc_cont}): discrete FID drops from 29.53 to 16.08 ($-$45\%) and continuous from 15.18 to 11.43 ($-$25\%), and the discrete--continuous gap shrinks from 14.35 to 4.65 FID.

\begin{table}[h]
\caption{\textbf{Controlled discrete vs.\ continuous comparison} (FID-50K$\downarrow$, ImageNet $256{\times}256$, same Transfusion framework).}
\label{tab:disc_cont}
\centering
\small
\begin{tabular}{ccccc}
\toprule
Step & Continuous & Continuous\textsubscript{\scriptsize +\method{}} & Discrete & Discrete\textsubscript{\scriptsize +\method{}} \\
\midrule
72k & 29.05 & --- & 40.03 & --- \\
96k & 22.07 & --- & 36.46 & --- \\
130k & 18.26 & --- & 32.98 & --- \\
\midrule
162k & 17.66 & 17.10 & 31.32 & 30.38 \\
184k & 16.72 & 14.02 & 29.58 & 19.59 \\
226k & 15.00 & 12.12 & 30.46 & 16.78 \\
266k & 15.18 & \textbf{11.43} & 29.53 & \textbf{16.08} \\
\bottomrule
\end{tabular}
\end{table}

\subsection{Additional text-to-image evaluations}
\label{app:t2i_extra}

Table~\ref{tab:t2i_extra}(a) reports GenAI-Bench~\cite{genaibench} VQAScore. Both models perform on par with T2I-R1~\cite{t2ir1} (basic/advanced overall: Internal\textsubscript{\scriptsize +\method{}} 0.89/0.72, External\textsubscript{\scriptsize +\method{}} 0.88/0.73, T2I-R1 0.88/0.73) and score above SD-XL~\cite{sdxl} and Midjourney v6. On DSG-1k~\cite{dsg}, Internal\textsubscript{\scriptsize +\method{}} and External\textsubscript{\scriptsize +\method{}} score 77.8 and 78.1 overall. Table~\ref{tab:t2i_extra}(b) repeats GenEval with 3 seeds. Table~\ref{tab:aesthetic} reports the LAION aesthetic score~\cite{laion5b} on TextAtlas. \method{} does not trade visual quality for textual fidelity, as the aesthetic score improves on all three subsets (\textit{e.g.}, 4.08$\to$4.37 on StyleTextSynth). The slightly lower CLIP score of Show-o2\textsubscript{\scriptsize +\method{}} in Table~\ref{tab:textatlas} is likely an artifact of CLIP's 77-token text limit, which truncates the long TextAtlas prompts and is insensitive to whether the rendered text is correct.

\begin{table}[h]
\caption{\textbf{Additional T2I benchmarks.} (a) GenAI-Bench~\cite{genaibench} VQAScore, with baseline rows from the GenAI-Bench paper and T2I-R1~\cite{t2ir1}. (b) GenEval with 3 seeds (mean$\pm$std).}
\label{tab:t2i_extra}
\centering
\small
\textit{(a) GenAI-Bench}\\[2pt]
\setlength{\tabcolsep}{3pt}
\begin{tabular}{lcccccccccccc}
\toprule
& \multicolumn{6}{c}{Basic} & \multicolumn{6}{c}{Advanced} \\
\cmidrule(lr){2-7} \cmidrule(lr){8-13}
Method & Attr. & Scene & Spat. & Act. & Part & Overall & Count & Diff. & Comp. & Neg. & Univ. & Overall \\
\midrule
SD-XL~\cite{sdxl} & 0.84 & 0.84 & 0.82 & 0.83 & 0.89 & 0.83 & 0.71 & 0.73 & 0.69 & 0.50 & 0.66 & 0.63 \\
Midjourney v6 & 0.88 & 0.87 & 0.87 & 0.87 & 0.91 & 0.87 & 0.78 & 0.78 & 0.79 & 0.50 & 0.76 & 0.69 \\
T2I-R1~\cite{t2ir1} & 0.87 & 0.89 & 0.89 & 0.87 & 0.87 & 0.88 & 0.81 & 0.82 & 0.78 & 0.60 & 0.73 & 0.73 \\
\midrule
Internal\textsubscript{\scriptsize +\method{}} & 0.89 & 0.91 & 0.90 & 0.89 & 0.88 & 0.89 & 0.79 & 0.80 & 0.77 & 0.58 & 0.76 & 0.72 \\
External\textsubscript{\scriptsize +\method{}} & 0.88 & 0.89 & 0.89 & 0.87 & 0.88 & 0.88 & 0.79 & 0.81 & 0.77 & 0.59 & 0.76 & 0.73 \\
\bottomrule
\end{tabular}

\vspace{6pt}
\textit{(b) GenEval, 3 seeds}\\[2pt]
\setlength{\tabcolsep}{3pt}
\begin{tabular}{lccccccc}
\toprule
Method & Single & Two & Count & Color & Pos. & C.Attr. & Overall \\
\midrule
Internal\textsubscript{\scriptsize +\method{}} & 0.98$\pm$0.01 & 0.91$\pm$0.02 & 0.68$\pm$0.02 & 0.86$\pm$0.02 & 0.76$\pm$0.02 & 0.55$\pm$0.03 & 0.79$\pm$0.01 \\
External\textsubscript{\scriptsize +\method{}} & 0.97$\pm$0.01 & 0.89$\pm$0.02 & 0.67$\pm$0.03 & 0.85$\pm$0.02 & 0.74$\pm$0.03 & 0.66$\pm$0.03 & 0.80$\pm$0.01 \\
\bottomrule
\end{tabular}
\end{table}

\begin{table}[h]
\caption{\textbf{Aesthetic quality on TextAtlas} (LAION aesthetic score~\cite{laion5b}$\uparrow$).}
\label{tab:aesthetic}
\centering
\small
\begin{tabular}{lccc}
\toprule
Method & StyleTextSynth & TextScenesHQ & TextVisionBlend \\
\midrule
Ground truth & 4.72 & 4.72 & 3.87 \\
Show-o2$^*$ & 4.08 & 4.01 & 3.40 \\
\rowcolor{tabhighlight} \textbf{Show-o2\textsubscript{\scriptsize +\method{}} (ours)} & \textbf{4.37} & \textbf{4.23} & \textbf{3.70} \\
\bottomrule
\end{tabular}
\end{table}

\subsection{Additional editing evaluation}
\label{app:edit_extra}

PIE-Bench focuses on preservation, so we also evaluate on I2EBench~\cite{i2ebench} with the original instructions (Table~\ref{tab:i2ebench}). Internal\textsubscript{\scriptsize +\method{}} is better on all low-level dimensions and comparable on high-level ones, consistent with an advantage specific to fine-grained detail preservation.

\begin{table}[h]
\caption{\textbf{I2EBench results} (higher is better).}
\label{tab:i2ebench}
\centering
\small
\begin{minipage}[t]{0.48\textwidth}
\centering
\textit{(a) Low-level editing}\\[2pt]
\begin{tabular}{lcc}
\toprule
Task & Internal & External \\
\midrule
Deblurring & \textbf{0.345} & 0.333 \\
HazeRemoval & \textbf{0.537} & 0.438 \\
Lowlight & \textbf{0.240} & 0.107 \\
RainRemoval & \textbf{0.583} & 0.460 \\
ShadowRemoval & \textbf{0.362} & 0.353 \\
SnowRemoval & \textbf{0.449} & 0.439 \\
WatermarkRemoval & \textbf{0.499} & 0.424 \\
RegionAccuracy & \textbf{0.536} & 0.509 \\
\bottomrule
\end{tabular}
\end{minipage}%
\hfill
\begin{minipage}[t]{0.48\textwidth}
\centering
\textit{(b) High-level editing}\\[2pt]
\begin{tabular}{lcc}
\toprule
Task & Internal & External \\
\midrule
Counting & \textbf{0.130} & 0.082 \\
DirectionPerception & \textbf{0.471} & 0.434 \\
ObjectRemoval & 0.446 & 0.446 \\
Replacement & 0.294 & \textbf{0.331} \\
BGReplacement & 0.496 & \textbf{0.525} \\
ColorAlteration & 0.246 & \textbf{0.254} \\
\bottomrule
\end{tabular}
\end{minipage}
\end{table}

\section{Limitations}
\label{app:limitations}

\paragraph{Scale and design choices.} \method{} has not been validated on larger-scale models, and scaling behavior remains future work. We study only SigLIP 2 as the visual semantic representation, and other semantic representations (CLIP, DINOv2) may offer different trade-offs. The progressive training schedule introduces hyperparameters ($p_0, p_1, k, \lambda_0, \lambda_1$) that we have not extensively tuned. Better schedules may yield further improvements.

\paragraph{Scope of the comparisons.} Our C2I experiments validate \method{} within matched pairs in different frameworks. Only the experiment in Appendix~\ref{app:disc_cont} compares discrete and continuous tokens under a single framework. The internal-vs-external comparison covers the evaluated configurations: the editing ablations (Table~\ref{tab:editing}b) show that the preservation gap stems from semantic-only DiT conditioning, and other external designs that inject reference features can largely close it. Our T2I and editing evaluations cover GenEval, GenAI-Bench, DSG-1k, TextAtlas, PIE-Bench, and I2EBench, but not human studies.

\paragraph{Failure cases.} We observe two failure modes. (1) When the generated SigLIP 2 tokens deviate from the intended semantics (\textit{e.g.}, rare categories with limited training data), the rendering stage inherits the error. (2) In editing, complex spatial rearrangements (\textit{e.g.}, ``move the cat's right leg onto its left leg'') remain challenging, as SigLIP 2 tokens capture semantic content rather than precise spatial transformations.

\section{Broader Impact}
\label{app:impact}

This work improves the quality and efficiency of image generation models, which carries both positive and negative societal implications. On the positive side, more efficient training reduces computational cost and energy consumption, making image generation research more accessible. On the negative side, higher-quality image generation could be misused to create misleading or harmful visual content such as deepfakes. We note that \method{} is a general training technique applied to existing open-source architectures and does not introduce new risks beyond those already present in the underlying models. We encourage the community to develop and deploy appropriate safeguards, including watermarking and content provenance tools, alongside advances in generation quality.

\end{document}